# آشکارسازی حالات لبخند و خندهٔ چهرهٔ افراد بر پایهٔ نقاط کلیدی محلی کمینه

**مینا محمدی دشتی[1] و مجید هارونی* [2]**

[1] دانشکده مهندسی کامپیوتر، واحد نجف آباد، دانشگاه آزاد اسلامی، نجف آباد، ایران

[2] گروه مهندسی کامپیوتر، واحد دولت آباد، دانشگاه آزاد اسلامی، اصفهان، ایران

## چکیده

**در این مقاله، آشکارسازی حالات لبخند و خنده چهره با رویکرد توصیف و کاهش بُعد نقاط کلیدی ارائه شده‌است. اساس کار در این پژوهش بر مبنای دو هدف استخراج نقاط محلی کلیدی و ویژگی ظاهری آن‌ها، و همچنین کاهش وابستگی سامانه به آموزش نهاده شده‌است. برای تحقق این اهداف سه سناریوی مختلف استخراج ویژگی ارائه شده است. ابتدا اجزای یک صورت توسط الگوریتم الگوی دودویی محلی آشکار می‌شود؛ سپس در سناریوی نخست، با توجه به تغییرات همبستگی پیکسل‌های مجاور بافت محدودهٔ لب، مجموعه نقاط کلیدی محلی بر پایهٔ گوشه‌یاب هریس استخراج می‌شود. در سناریوی دوم، کاهش بعد نقاط مستخرج سناریوی نخست با بهبود الگوریتم تحلیل مؤلفه‌های اصلی انجام می‌شود؛ و در سناریوی آخر با مقایسه مختصات نقاط مستخرج از سناریوی نخست و توصیف‌گر بریسک مجموعه نقاط بحرانی استخراج می شود. در ادامه بدون آموزش سامانه، با مقایسه شکل و فاصله هندسی نقاط محلی محدوده لب حالات چهره آشکار می شود. برای ارزیابی روش پیشنهادی، از پایگاه داده‌های استاندارد و شناخته‌شده Cohn-Kaonde، CAFE، JAFFE و Yale استفاده شده‌است. نتایج به‌دست‌آمده از سناریوهای مختلف به‌ترتیب بیان‌گر بهبود ۶/۳۳ و ۱۶/۴۶ در صدی متوسط نرخ دقت بازشناسی سناریوی دوم نسبت به نخست و سناریوی سوم نسبت به دوم است. همچنین نتایج کلی آزمایش‌ها، کارایی قابل قبول بالای ۹۰ در صد روش پیشنهادی را نشان می‌دهد.**

# Smile and Laugh Expressions Detection Based on Local Minimum Key Points

**Mina Mohammadi Dashti[1] & Majid Harouni[2*]**

[1]Faculty of Computer Engineering, Najafabad Branch, Islamic Azad University, Najafabad, Iran

[2]Department of Computer Engineering, Dolatabad Branch, Islamic Azad University, Isfahan, Iran

## Abstract

**In this paper, a smile and laugh facial expression is presented based on dimension reduction and description process of the key points. The paper has two main objectives; the first is to extract the local critical points in terms of their apparent features, and the second is to reduce the system's dependence on training inputs. To achieve these objectives, three different scenarios on extracting the features are proposed. First of all, the discrete parts of a face are detected by local binary pattern method that is used to extract a set of global feature vectors for texture classification considering various regions of an input-image face. Then, in the first scenario and with respect to the correlation changes of adjacent pixels on the texture of a mouth area, a set of local key points are extracted using the Harris corner detector. In the second scenario, the dimension**

* Corresponding author

* نویسندهٔ عهده‌دار مکاتبات

reduction of the extracted points of first scenario provided by principal component analysis algorithm leading to reduction in computational costs and overall complexity without loss of performance and flexibility; and in the final scenario, a set of critical points is extracted through comparing the extracted points' coordinates of the first scenario and the BRISK Descriptor, which is utilized a neighborhood sampling strategy of directions for a key-point. In the following, without training the system, facial expressions are detected by comparing the shape and the geometric distance of the extracted local points of the mouth area. The well-known standard Cohn-Kaonde, CAFÉ, JAFFE and Yale benchmark dataset are applied to evaluate the proposed approach. The results shows an overall enhancement of 6.33% and 16.46% for second scenario compared with first scenario and third scenario compared with second scenario. The experimental results indicate the power efficiency of the proposed approach in recognizing images more than 90 % across all the datasets.

**Keywords: Local key points extraction, facial expression detection, corner detector, descriptor algorithm, dimension reduction**

## ۱- مقدمه

حالات چهره انسان ویژگی‌های منحصربه‌فردی را می‌توانند در بر داشته باشند. ازاین‌رو آشکارسازی حالت چهره مبین اطلاعات ظاهری است و به تبع آن کمک به تشخیص تغییرات رفتاری، روحی و روانی آنی یک شخص می‌کند. این موضوع نقش به‌سزای در گسترش ادراک فعالیت‌های شناختی و بهبود ارتباطات غیر کلامی یک شخص و یا بیمار ایفا می‌کند [1]. عوامل مؤثری همچون متنوع بودن، تغییرپذیری ژست و ظاهر، افزایش سن و گرفتگی چهره از چالش‌های آشکارسازی حالت‌های آن محسوب می‌شوند. فرایند سامانه آشکارسازی در چهار مرحلهٔ کلی، پیش‌پردازش، قطعه‌بندی، استخراج ویژگی و طبقه‌بندی‌کننده انجام می‌گیرد [2]. بسته به روش پژوهش پیشنهادی ترکیب یا ادغام این مراحل و یا تمرکز بر روی یک مرحله خاص را مد نظر می‌توان قرار داد. مرحله استخراج ویژگی به‌عنوان چالش مستقیم به‌سازی این نوع سامانه‌ها مورد توجه پژوهش‌گران قرار گرفته است که می‌بایست دارای دو شرط پایه ثبات اطلاعات در تصاویر مختلف و قدرت تمایز باشد [3,4]. در یک نگاه کلی، ویژگی‌های یک شیء تصویر می‌توانند از نوع آماری، مانند تراکم پیکسل‌ها، ساختاری، مانند خط و کمان، و یا ترکیبی از این دو نوع باشند [5]؛ همچنین در یک دسته‌بندی دیگر، ویژگی‌های یک تصویر به دو نوع محلی یا سراسری می‌توانند تقسیم‌بندی شوند [6-8]. در این میان ویژگی‌های محلی حالات چهره به استخراج نقاط کلیدی و الگوهای آماری بافت تصویر اطلاق می‌شوند. نحوه عملکرد و استخراج این دو گونه ویژگی‌های محلی چالش‌های جدی جهت کارایی سامانهٔ آشکارسازی حالت چهره و انطباق خصوصیات منحصر به‌فرد هر جزء از چهره را ایجاد کرده‌اند. عواملی که استخراج این‌گونه ویژگی‌ها را تحت تأثیر خود قرار می‌دهند، شامل مواردی همچون وجود اختلاف زیاد چهرهٔ افراد بر اساس نژاد، منطقهٔ جغرافیایی، و رنگ پوست هستند که سبب می‌شوند چهره‌ها را در دسته‌های مشخص‌شده‌ای نتوان طبقه‌بندی کرد [9]، تغییرات در شدت نور و یا تغییرات ظاهری مثل کوتاهی یا بلندی موی سر و چهره یا نحوه مرتب‌کردن آن‌ها و بستن چشم‌ها، گذاشتن عینک، سبیل و ریش، پوشش شال و کلاه، و حتی تغییر سن ممکن است، باعث تغییر بافت تصویر و یا چهره شوند [10]. موارد ذکر شده علاوه بر ایجاد مشکلات کلیدی برای استخراج ویژگی‌های محلی برای تشخیص و آشکارسازی چهره نیز موانعی را به‌وجود می‌آورند. از طرفی دیگر، جهت تشخیص حالات چهره بر پایهٔ ویژگی‌های محلی می‌بایست ابتدا محدودهٔ چهره در هر تصویر ورودی آشکار شود. روش‌های آشکارسازی چهره به دو روش مبتنی بر ظاهر و مبتنی بر ویژگی دسته‌بندی می‌شوند [11]. در روش‌های مبتنی بر ظاهر، هدف یافتن یک توصیف کلی از چهره است، استفاده از الگوریتم‌های تحلیل مؤلفه‌های اصلی (PCA)[1] توسط [12] در سال ۱۹۹۱ و در ادامه توسط [13] در سال ۲۰۱۴، تحلیل تفکیک خطی (LDA)[2] [14,15]، و تحلیل مؤلفه‌های مستقل(ICA)[3] [16,17]، در این روش‌ها پیشنهاد شده است. در روش‌های مبتنی بر ویژگی، هدف استخراج ویژگی‌های منحصربه‌فرد جهت تعیین نقاط برجسته یک تصویر است. در این میان روش‌های بر پایه الگوریتم‌های آشکارساز لبه [18]، خطوط راست [19,20]، و گوشه [21-23] پیشنهاد شده است؛ که با توجه به نوع بافت چهره و اجزای آن، استخراج و آشکارسازی گوشه‌های تصویر علاوه بر تفکیک‌پذیری و تمایز بین حالت چهره، از ثبات اطلاعات لازم نسبت به دو نوع دیگر می‌تواند برخوردار باشد. بازشناسی حالات چهره بر اساس استخراج نقاط کلیدی مبتنی بر الگوریتم گوشه‌یاب (FAST)[4] توسط *گاوو* و همکاران در سال

---
[1] Principal Component Analysis
[2] Linear Discriminant Analysis
[3] Independent Component Analysis
[4] Features from Accelerated Segment Test

۲۰۱۵ انجام شد که نقاط گوشه بلااستفاده توسط یک ماسک صورت، فیلتر و حذف شد [24]. لبه‌های تصویر توسط دو طبقه‌بندی‌کننده فاصله اقلیدسی آشکار شده سپس حالات چهره بر پایه شبکه عصبی بازشناسی می‌شود [25] و یا ترکیبی از الگوریتم‌های لبه‌یابی جهت آشکارسازی حالات چهره پیشنهاد شده است [26,27]. در تمامی روش‌های ارائه‌شده مبنا بر استخراج ویژگی‌های سراسری چهره بوده است که نیاز سامانه به آموزش مشهود است؛ مانند [28] که پس از استخراج ویژگی گابور، یک مدل پیش‌گوی مبتنی بر منیفولد محلی خطی پیشنهاد شده است؛ در صورتی که در بازشناسی حالات بر پایهٔ ویژگی‌های محلی، علاوه‌بر کاهش محاسبات پردازش، از وابستگی ذاتی نقاط کلیدی مجاور نیز می‌توان بهره برد؛ با این عمل از خصوصیت ظاهری و هندسی این نقاط نسبت به هم استفاده شده و در نتیجه مرحله آموزش سامانه می‌تواند حذف شود.

هدف اصلی این مقاله ارائه یک روش ترکیبی آشکارسازی حالت خنده و لبخند چهرهٔ افراد بر پایه ویژگی‌های محلی و ظاهری تصویر است. در این راستا، افزایش توان تعمیم آشکارسازی حالت چهره به‌کمک افزایش توان محاسباتی الگوریتم‌ها در سه سناریو مورد توجه قرار گرفته است. در سناریوی نخست، ویژگی‌های مکانی و ساختاری تصاویر بر اساس تغییرات همبستگی پیکسل‌های مجاور بافت استخرج می‌شود؛ بدین صورت که، از الگوریتم گوشه‌یاب هریس (Harris)[1] جهت استخراج ویژگی با خصوصیت اصلی حساسیت به چرخش تصویر استفاده شده است. در سناریوی دوم، از الگوریتم ترکیبی PCA و Harris جهت کاهش طول بردار ویژگی استفاده شد؛ که علاوه‌بر حذف ویژگی‌های نامفید، باعث کاهش زمان محاسبات و افزایش دقت بازشناسی می‌شود؛ سناریوی سوم بر اساس ترکیب دو الگوریتم Harris و نقاط کلیدی مقیاس‌پذیر ثابت دودویی مقاوم (BRISK)[2] است که مقایسهٔ توصیفی ضرایب نقاط کلیدی الگوریتم BRISK با الگوریتم Harris در نرخ دقت بازشناسی مورد بررسی قرار گرفته است. مزیت اصلی این سناریو کاهش حساسیت سامانه به تغییرات چرخش و مقیاس تصویر است. در روش پیشنهادی این مقاله، ضمن بهره‌گیری از مفاهیم به‌کار رفته در این سه سناریو، آشکارسازی حالات خنده و لبخند چهره بدون آموزش و مبتنی بر شکل، مکان هندسی و فاصلهٔ اقلیدسی نقاط محلی از هم‌دیگر ارائه شده است.

در ادامه این مقاله، در بخش دوم، پایگاه‌داده‌های شاخص جهت ارزیابی روش پیشنهادی بررسی می‌شود. در بخش سوم روش پیشنهادی به‌تفصیل توضیح داده شده است. در بخش بعدی نتایج پیاده‌سازی، آزمایش‌های انجام‌شده و ارزیابی بیان شده است. و بخش آخر به نتیجه‌گیری اختصاص داده شده است.

## ۲- پایگاه‌داده‌های شاخص

در این پژوهش از تصاویر سه حالت مختلف چهره، یعنی معمولی، لبخند و خنده، پایگاه‌داده‌های استاندارد Cohn-Kaonde، JAFFE، CAFE و Yale استفاده شده است. پایگاه‌داده Cohn-Kaonde شامل ۱۴۳ تصویر بین بازهٔ سنی ۳۰-۱۸ سال بوده که از چندین کشور مختلف با دو جنسیت متفاوت، دارای کلاه، رنگ پوست متفاوت، مدل موهای مختلف هستند [29]. پایگاه‌داده JAFFE شامل ۲۱۳ تصویر سیاه و سفید از هفت حالت چهرهٔ ده خانم جوان ژاپنی است که در دانشگاه کیوشوی ژاپن تهیه شده است [30]. پایگاه داده CAFE شامل ۱۳۹ تصویر سیاه و سفید از شش حالت چهره در دانشگاه کالیفرنیا است. این تصاویر از حالت چهرهٔ متفاوت خشم، نفرت، تعجب، شاد، غم‌انگیز، ترس و تعجب از ۲۴ نفر متفاوت در سنین مختلف جمع‌آوری شده‌اند [31]. پایگاه داده Yale شامل ۱۶۵ تصویر سیاه و سفید از پانزده نفر با سنین مختلف و هر دو جنسیت تهیه شده است؛ برای هر نفر یازده تصویر متفاوت در حالت‌های همچون حالت غمگین، تعجب و شادی ضبط شده است [32]. خلاصهٔ مشخصهٔ پایگاه‌داده‌ها در جدول (۱) بیان و نمونه‌ای از تصاویر آن‌ها در شکل (۱) نشان داده شده است.

**(جدول-۱): ویژگی های پایگاه‌داده‌های شاخص**
**(Table-1): Properties of the benchmark datasets**

| نام | تعداد تصاویر | وضوح تصویر | توضیح |
|---|---|---|---|
| Cohen-kanade | 486 | 490×640 | کلاه و بدون کلاه، موهای بلند و کوتاه رنگ پوست سیاه و سفید، هر دو جنسیت زن و مرد، |
| JAFFE | 213 | 256×256 | سیاه و سفید، پس‌زمینه تیره و روشن، جنسیت زن |
| CAFE | 139 | 320×243 | موهای بلند و کوتاه، پس‌زمینه تیره و روشن، هر دو جنسیت زن و مرد |
| Yale | 165 | 320×243 | عینک و ریش و سبیل، نورهای مختلف، هر دو جنسیت زن و مرد |

[1] Harris Corner Detector
[2] Binary Robust Invariant Scalable Key-points

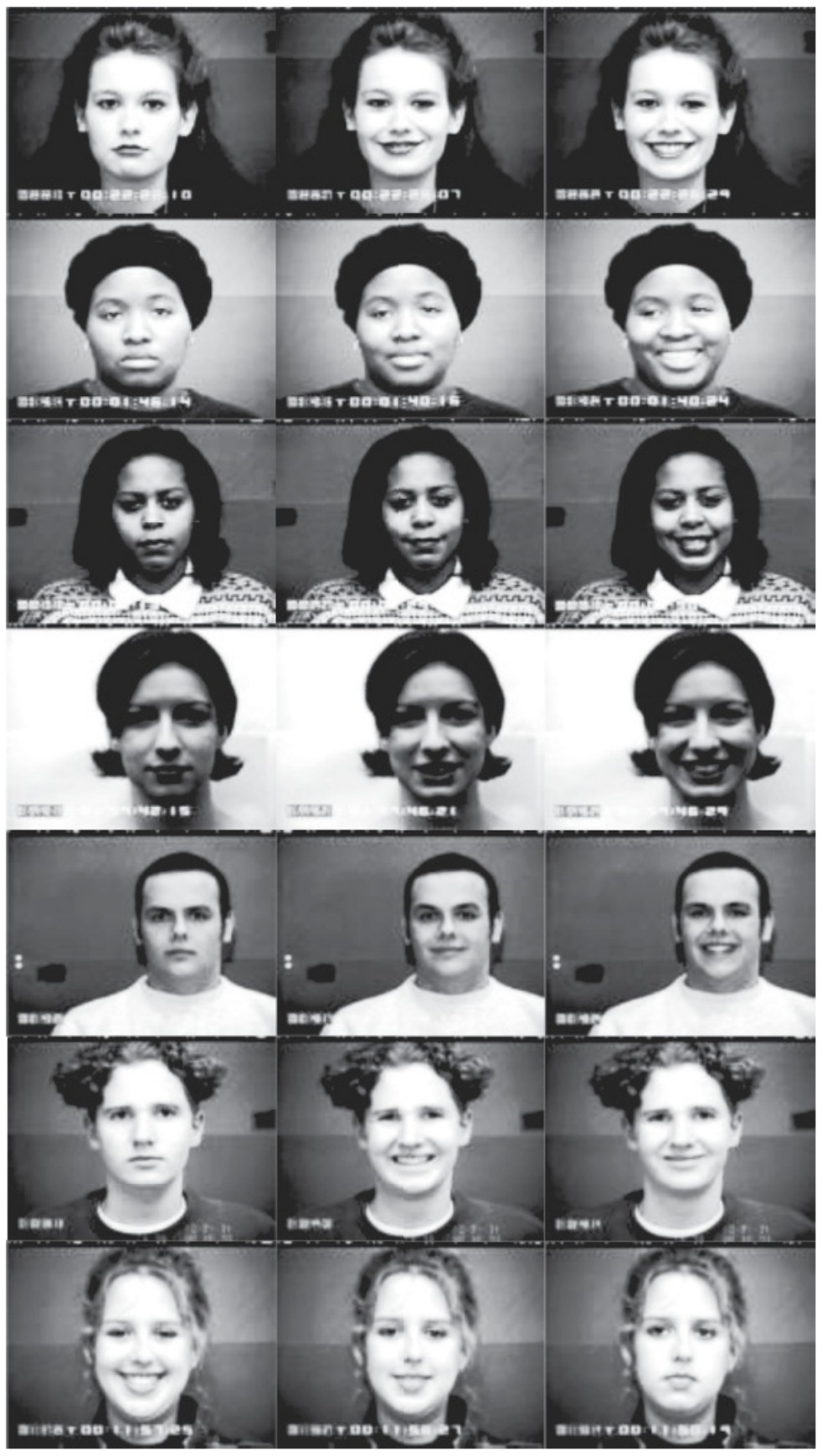

(الف)

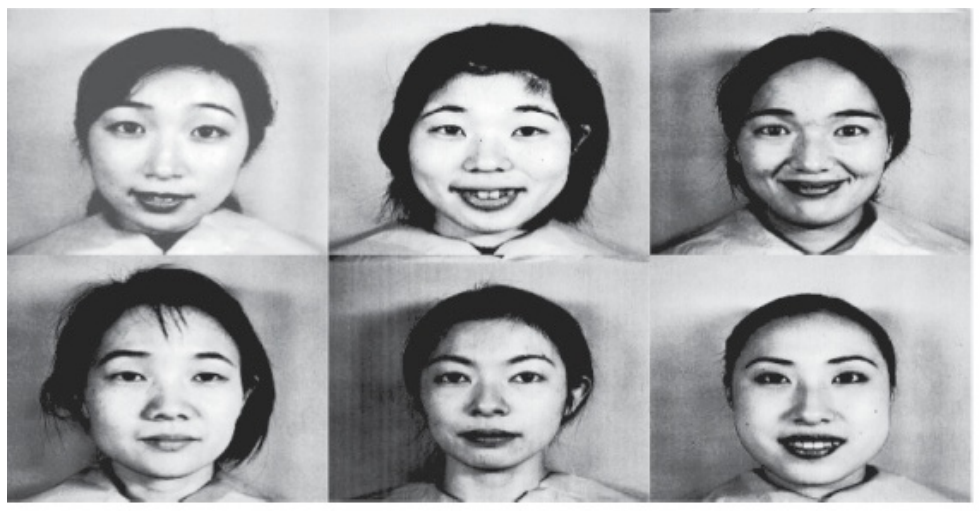

(ب)

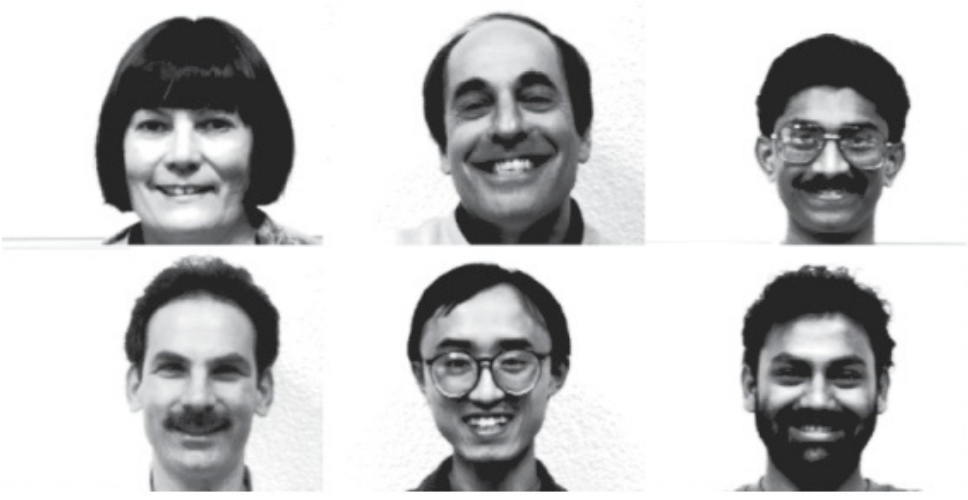

(ج)

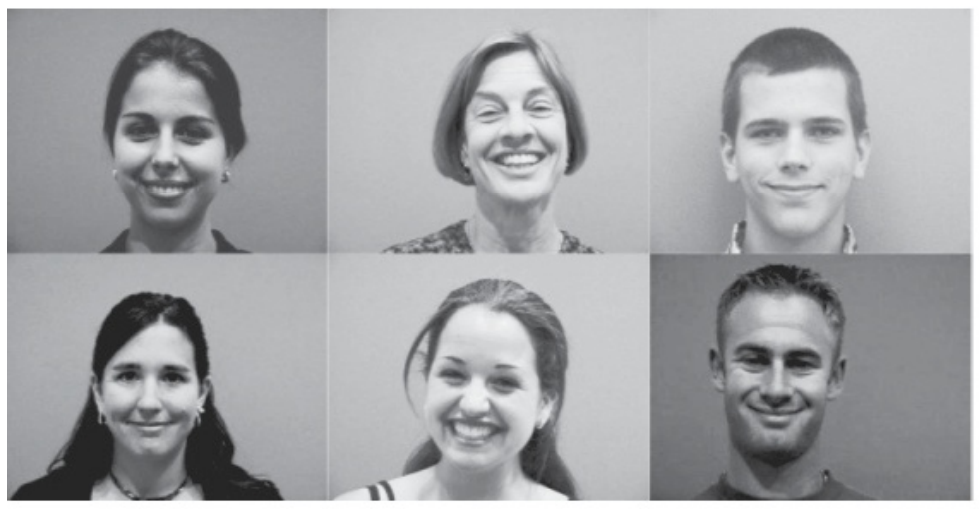

(د)

**(شکل-۱): نمونه‌های از تصاویر پایگاه داده‌ها: الف) Cohn-Kaonde، ب) JAFFE، ج) Yale، د) CAFE**

**(Figure-1): Sample facial expression images in datasets: a) Cohn-Kaonde, b) JAFFE, c) Yale, d) CAFE**

## ۳- روش پیشنهادی

شکل (۲) چارچوب کلی روش پیشنهادشده در این پژوهش را نشان می‌دهد. ابتدا با دریافت تصاویر ورودی، عملیات پیش‌پردازش تبدیلات سطح خاکستری و شدت روشنایی اعمال می‌شود تا تصاویر مطلوب‌تری حاصل شود؛ مطالعات نشان می‌دهد که نتایج مرحلهٔ پیش‌پردازش تأثیر مستقیمی در کارایی مطلوب و سرعت یک سامانه می‌تواند داشته باشد؛ در ادامه، آشکارسازی ناحیهٔ صورت و همچنین محدوده لب بر پایه روش الگوی دودویی محلی (LBP)[1] پیاده‌سازی شد؛ سپس جهت بازشناسی حالات لبخند و خنده چهره سه روش جداگانه مبتنی بر گوشه‌یابی، کاهش ابعاد داده و استخراج نقاط کلیدی تصاویر ارائه شده است؛ که بر پایه نمایش اشکال هندسی نقاط مستخرج کلیدی و خصوصیات منحنی بزیر و یا توابع برازش درجهٔ دوم ایجادشده از مختصات این نقاط و همچنین فاصله اقلیدسی میان نقاط بیشینه و کمینه این توابع، بازشناسی نهایی حالات چهره انجام می‌شود. ازآنجایی‌که این تعمیم بازشناسی با تمرکز بر افزایش کارایی روش پیشنهادی و کاهش ابعاد نقاط مستخرج کلیدی و وابستگی به یک طبقه‌بند شبیه‌سازی شده است، مبنای کار بر پایه حذف نقاط با اطلاعات ناکافی و مشابه و یا با اطلاعات اشتباهی است که به‌طور مستقیم بر افزایش بار محاسباتی سامانه تأثیر دارند و یا حتی ممکن است، سبب تولید خروجی نامطلوب نیز شوند؛ بنابراین در بازشناسی حالات چهره بدون آموزش، بررسی سناریوهای پالایش این نقاط کلیدی، امری ضروری به نظر می‌رسد. ازهمین‌رو، اساس کار، انتخاب استخراج نقاط کلیدی یک حالت چهره خواهد بود که به‌تنهایی در توصیف آن حالت مفید باشد.

[1] Local Binary Pattern

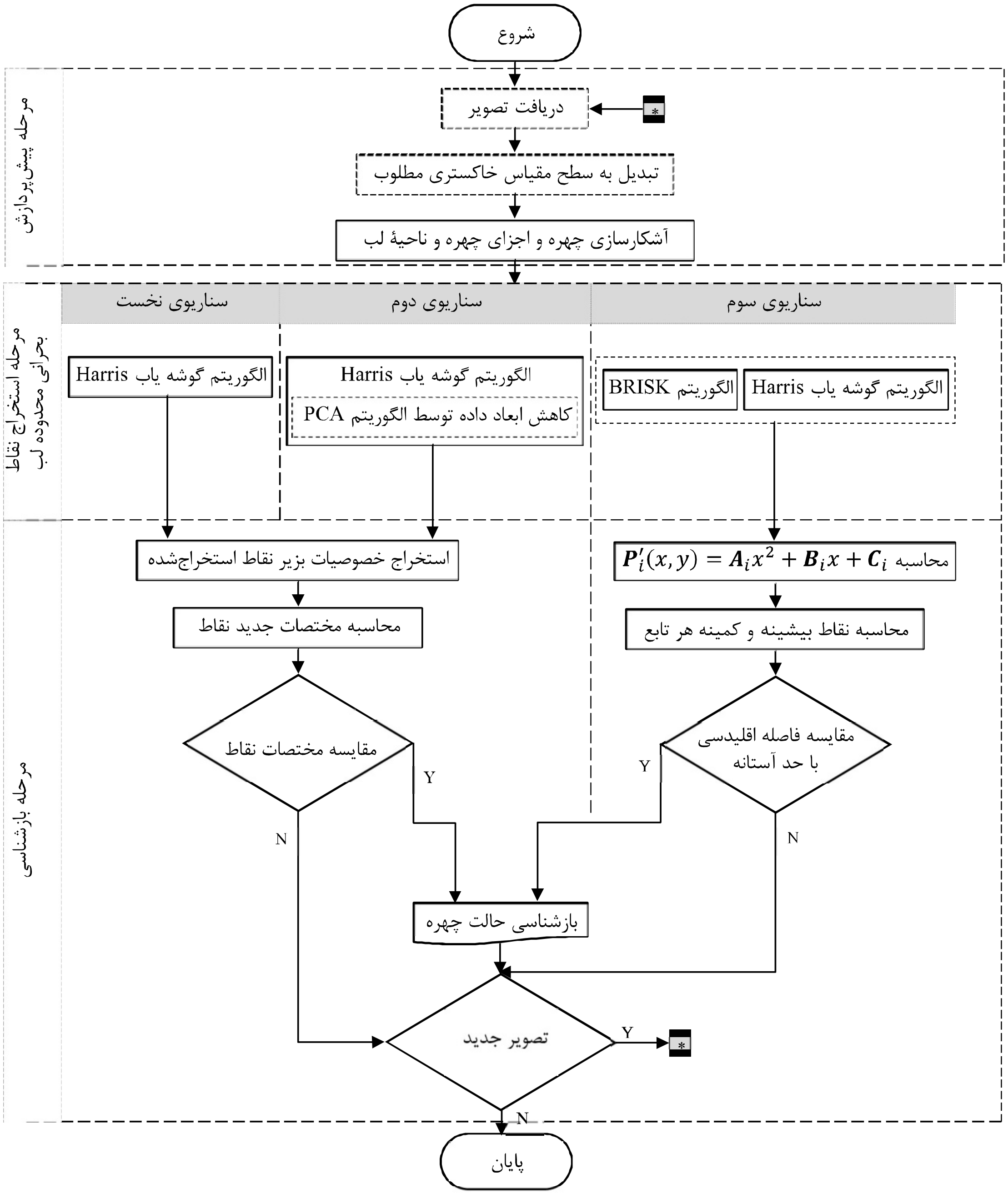

**(شکل-۲): چارچوب روند کلی الگوریتم پیشنهادی**

**(Figure-2): Flow-chart corresponding to the proposed algorithm.**

## ۱-۳- پیش‌پردازش تصاویر و آشکارسازی محدوده‌های مورد نظر

در این بخش ابتدا تصاویر ورودی خام به تصاویر مطلوب تبدیل شده و سپس آشکارسازی چهره و محدودهٔ لب انجام می‌گیرد. جهت کاهش ابعاد پیکسل تصاویر خام ورودی و همچنین کاهش پیچیدگی محاسبات در مراحل بعدی سامانه، هر تصویر به سطح مقیاس خاکستری توسط رابطه (۱) تبدیل می‌شود.

$$G_L_{(x,y)} = 0.30 * R_{(x,y)} + 0.59 * G_{(x,y)} + 0.11 * B_{(x,y)} \quad (۱)$$

که $G_L$ مؤلفهٔ خروجی سطح خاکستری تصویر و $R$، $G$ و $B$ به‌ترتیب مؤلفه‌های قرمز، سبز، آبی هر پیکسل از

تصویر ورودی هستند [33]. در ادامه جهت تعدیل و بهبود شدت روشنایی تصاویر بالا از الگوریتم بهبودیافته تطبیقی معکوس تانژانت هذلولوی (E-AIHT)[1] که بر پایه الگوریتمی است که در منبع [**9**] در سال ۲۰۰۱ است، استفاده می‌شود. در این الگوریتم بهبود شدت روشنایی به‌صورت رابطه (۲) انجام می‌شود.

$$G_L_{(i,j)}=\left(\log\left(\frac{1+G_L_{(i,j)}^{Bias\left(G_L_{(x,y)}\right)}}{1-G_L_{(i,j)}^{Bias\left(G_L_{(x,y)}\right)}}\right)-1\right)\times Gain\left(G_L_{(x,y)}\right) \quad (۲)$$

که $G_L_{(i,j)}$ مقدار سطح خاکستری در سطر $i$ام و ستون $j$ام تصویر است. همچنین توابع کنترل $Bias\left(G_L_{(x,y)}\right)$ و $Gain\left(G_L_{(x,y)}\right)$ به‌ترتیب نشان‌دهنده سرعت تغییر روشنایی پیکسل و تابع وزنی شیب تند منحنی E-AIHT هستند که از رابطه‌های (۳)، (۴) و (۵) قابل محاسبه هستند.

$$Bias\left(G_L_{(x,y)}\right)=\left(\frac{\frac{1}{m\times n}\sum_{i=1}^{m}\sum_{j=1}^{n}G_L_{(x,y)}}{\alpha}\right)^{\beta} \quad (۳)$$

$$\begin{aligned}Gain\left(G_L_{(x,y)}\right)&=\rho\times\left(variance\right)^{\gamma}\\&=\rho\times\left(\frac{1}{m\times n}\sum_{i=1}^{m}\sum_{j=1}^{n}\left(G_L_{(x,y)}-\mu\right)\right)^{\gamma}\end{aligned} \quad (۴)$$

$$\mu=\frac{1}{m\times n}\sum_{i=1}^{m}\sum_{j=1}^{n}G_L_{(x,y)} \quad (۵)$$

که $\alpha$ و $\beta$ پارامترهای قابل تنظیم جهت مشخص‌کردن سرعت تغییرات روشنایی پیکسل‌ها، $\rho$ و $\gamma$ پارامترهای وزنی شیب تند منحنی و همچنین، $m$ و $n$ تعداد سطرها و ستون‌های تصویر هستند. با اعمال این الگوریتم تصاویر سطح خاکستری مطلوب $G_{Desired}^{L}$ به‌دست می‌آید.

در سامانه‌های بازشناسی حالات چهره، آشکارسازی چهره و محدودهٔ لب اهمیت زیادی دارند. در این میان، مشخص‌کردن محدودهٔ لب به‌منظور کاهش زمان پردازش و کاهش پیچیدگی محاسبات بر روی کل تصویر خواهد بود. در این پژوهش، روش تحلیل بافت LBP به دلیل سادگی محاسبات و همچنین پایداری در برابر تغییرات اندک روشنایی تصاویر ورودی به‌عنوان روش آشکارسازی محدوده‌های مورد نظر تصاویر به‌کار گرفته شده است [34]. این روش در دو مرحلهٔ کلی ابتدا جهت جداسازی محدودهٔ صورت و سپس تخمین محدودهٔ لب بر روی آن پیاده‌سازی شده است. بر پایه منبع [35] الگوریتم وایولا-جونز برای استخراج محدوده‌های مورد نظر استفاده شده است. این الگوریتم ابتدا تصویر ورودی را به زیرتصاویر یا پنجره‌های به اندازه ۲۴×۲۴ تقسیم‌بندی کرده و سپس در چهار گام اصلی کار می‌کند: نخست، استخراج ویژگی‌های هار[2] (شکل (۳))؛ دوم، ارزیابی ویژگی‌ها توسط تصاویر انتگرال[3] (رابطه (۶) و شکل (۴))؛ سوم، استفاده از الگوریتم یادگیری آدابوست[4] جهت انتخاب بهترین ویژگی‌ها، در این مرحله ترکیب دسته‌بندهای ضعیف باعث به‌وجودآمدن یک دسته‌بندی قوی می‌شود؛ و در آخر استفاده از طبقه‌کننده آبشاری جهت تمایز بین تصاویر صورت/محدودهٔ لب و یا غیر صورت/محدودهٔ لب.

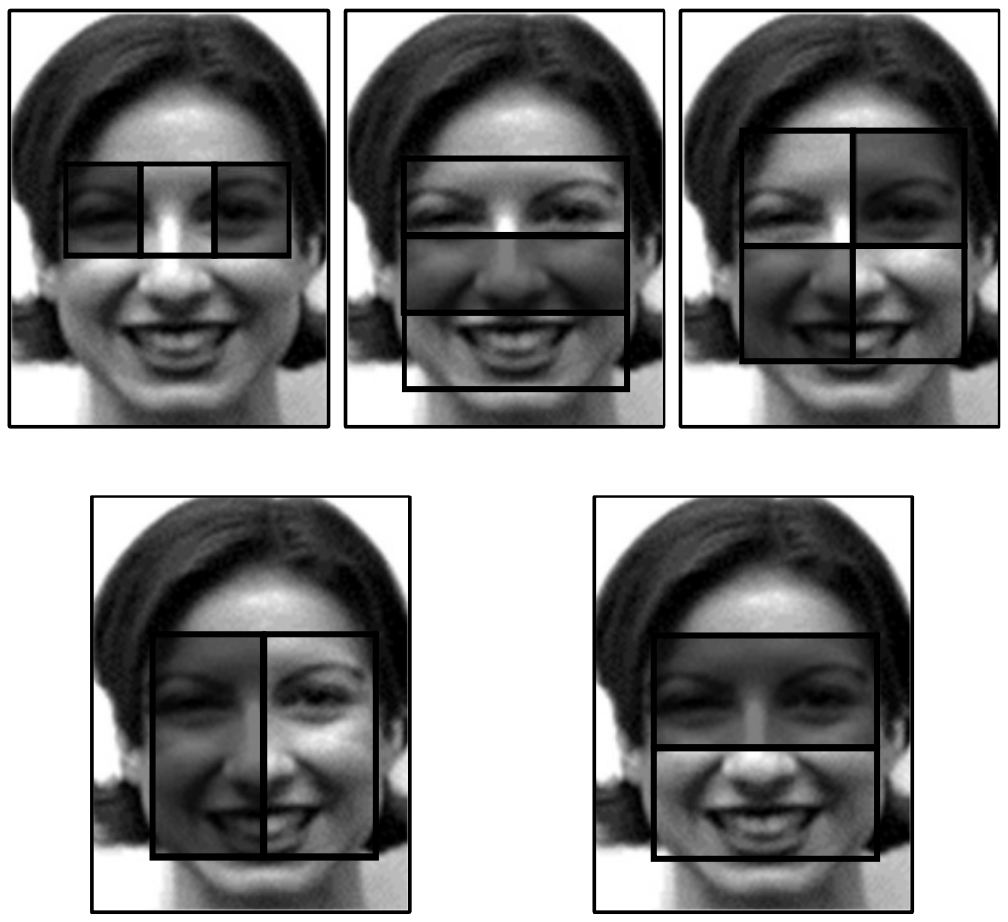

**(شکل-۳): شمای کلی از ویژگی‌های هار**
**(Figure-3): Overview of proposed Haar features set.**

$$I_G_{Desired}^{L}(x,y)=\sum_{x'\le x,\ y'\le y}G_{Desired}^{L}(x',y') \quad (۶)$$

که $I_G_{Desired}^{L}$ مجموع تصاویر انتگرالی و $G_{Desired}^{L}$ تصویر واقعی مطلوب پیش‌پردازش شده هستند.

[1] Enhanced Adaptive Inverse Hyperbolic Tangent Algorithm

[2] Haar Features
[3] Integral Images
[4] Ada-boost

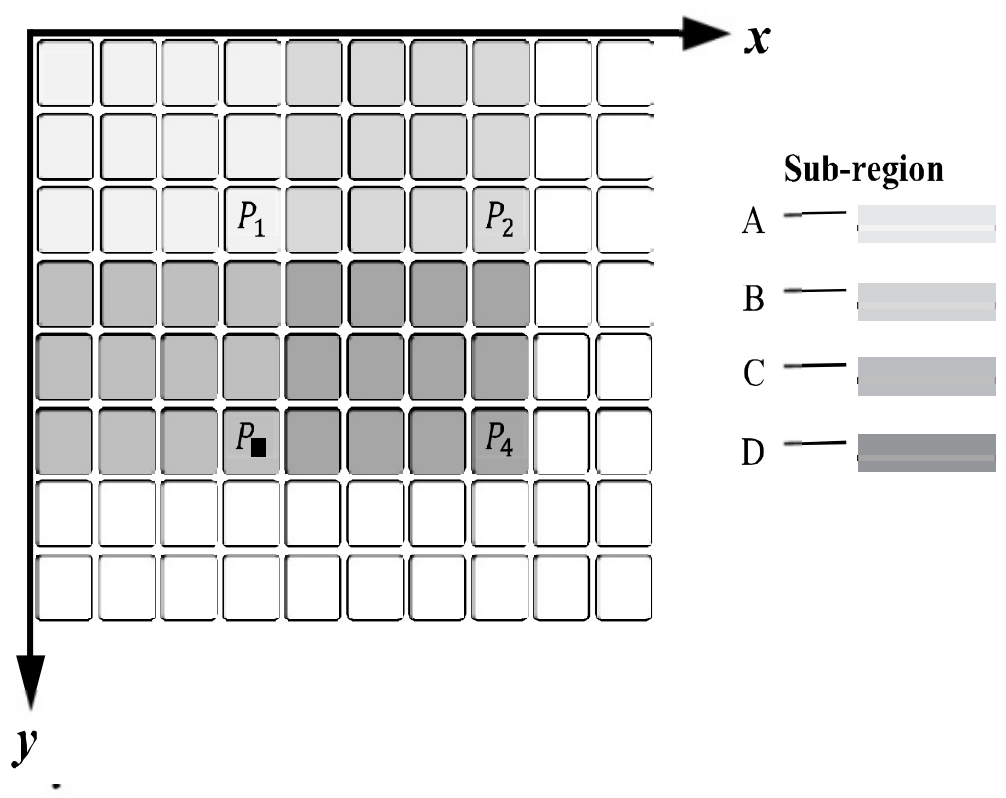

**(شکل-۴): مقادیر پیکسل‌های $P_1$، $P_2$، $P_3$ و $P_4$ در تصویر انتگرالی به ترتیب برابر با مجموع مقادیر پیکسل‌های محدوده A، مجموع مقادیر پیکسل‌های محدوده A+B، مجموع مقادیر پیکسل‌های محدوده A+C و مجموع مقادیر پیکسل‌های محدوده A+B+C+D در تصویر اصلی پیش‌پردازش شده هستند. مجموع مقادیر پیکسل‌ها در محدوده D برابر است با D−B−C+A.**

**(Figure-4): The pixels values of P1, P2, P3 and P4 in the integral images are, respectively, equal to the sum of all pixels values in A area, the sum of all pixels values in A+B area, the sum of all pixels values in A+C area, and the sum of all pixels values in A+B+C+D area in the pre-processed (original) image. The sum of all pixels values in D area is D−B−C+A.**

## ۲-۳- استخراج نقاط بحرانی محلی و بازشناسی حالات چهره

جهت استخراج نقاط بحرانی محدودهٔ لب و سپس محاسبات لازم جهت بازشناسی حالات خنده و لبخند چهره سه سناریوی مختلف پیشنهاد و مورد بررسی قرار گرفته است.

### ۱-۲-۳- سناریوی نخست: استخراج نقاط گوشه

در این روش ابتدا نقاط گوشه به‌عنوان نقاط بحرانی در محدودهٔ لب توسط الگوریتم Harris استخراج می‌شود [34]. برای این منظور، انتخاب پیکسل‌های همسایهٔ یک پیکسل مرکزی توسط یک پنجره گاوسی بر اساس رابطه (۷) به‌دست می‌آید.

$$I_w(x,y) = exp\left(-\frac{x^2+y^2}{2\sigma^2}\right) \qquad (۷)$$

که $\sigma$ اندازه پنجرهٔ گوسی است. جزییات بیشتر مربوط به نحوهٔ محاسبات و پیاده‌سازی الگوریتم Harris در [34,36] آورده شده است. با به‌دست‌آورن میزان جابه‌جایی یک پنجره تصویر در محورهای $x$ و $y$ و همچنین استفاده از یک ماتریس ۲×۲ از طریق رابطه (۸) میزان تغییر جهت‌ها و پراکندگی پیکسل‌ها را می‌توان محاسبه کرد.

$$D = \sum_{X,Y} I_w(x,y)\begin{bmatrix} I_X^2 & I_X I_Y \\ I_X I_Y & I_Y^2 \end{bmatrix} \qquad (۸)$$

جایی که $I_X$ و $I_Y$ میزان شدت پیکسل تصویر[1] در محورهای $x$ و $y$، و $I_w$ تصویر پنجره‌دار شده هستند؛ و رابطه‌های (۹)، (۱۰) و (۱۱) مشخص می‌کنند که وضعیت و ویژگی ظاهری پیکسل مرکزی چگونه است:

$$f_{Respond} = det(D) - kTr(D)^2 \qquad (۹)$$

$$\det(D) = I_X^2 I_Y^2 + (I_X I_Y)^2 \qquad (۱۰)$$

$$kTr(D) = I_X^2 + I_Y^2 \qquad (۱۱)$$

در رابطه‌های بالا $k$ یک پارامتر ثابت تجربی است که بین ۰/۰۴ و ۰/۰۶ در نظرگرفته می‌شود [37]، و $f_{Respond}$ تابع پاسخی است که اگر مقدار آن مثبت، منفی و یا خیلی کوچک‌تر از یک حد آستانه باشد، به‌ترتیب مشخص‌کنندهٔ این است که پیکسل مرکزی در ناحیهٔ گوشه، لبه و یا یک سطح صاف قرار دارد؛ بدین ترتیب پیکسل‌های گوشه در محدودهٔ لب به‌صورت رابطه (۱۲) ذخیره می‌شوند.

$$P^{II}_{corner}(x,y) = \{(x_1,y_1)\ldots(x_i,y_i) | 1 \le i \le n'\} \qquad (۱۲)$$

که در آن $n'$ تعداد نقاط گوشه در محدودهٔ لب است.

در ادامه، جهت بازشناسی حالت چهره از نمایش اشکال هندسی نقاط بالا توسط الگوریتم (۱) که مبتنی بر خصوصیات منحنی بزیر است، استفاده می‌شود. در تفسیر و بازشناسی حالت شکل نقاط به‌دست‌آمده ابتدا یک بردار با فضای خطی به اندازه $n'$ ایجاد و سپس ماتریس $n' \times n'$ ایجاد می‌شود که با مقادیر قابل محاسبه از موقعیت سطری و ستونی هر عنصر و همچنین مقادیر بردار ذکرشده مقداردهی می‌شود. در ادامه، مختصات جدید نقاط کلیدی بر پایه ترانهاده بردار و همچنین ماتریس ایجادشده، محاسبه و با مقایسه مختصات نقاط جدید و قبلی، حالت چهره بازشناسی می‌شود.

### ۲-۲-۳- سناریوی دوم: کاهش بُعد نقاط داده‌ها

در این روش پیکسل‌های گوشه در محدودهٔ لب مطابق روش پیشنهادی قبلی بر پایه الگوریتم Harris استخراج می‌شود، سپس در فضای برداری توسط الگوریتم PCA کاهش بعد مجموعه داده‌های این پیکسل‌ها انجام می‌گیرد. درواقع به‌کمک ماتریس کوواریانس که در رابطه (۱۳) نشان داده

[1] Image Pixel Intensity

شده‌است مقادیر ویژهٔ داده‌ها به یک زیرفضای جدید منتقل می‌شوند. به این ترتیب مؤلفه‌هایی از مجموعه‌داده که بیشترین تأثیر در واریانس را دارند، حفظ شده و مابقی حذف خواهند شد [38,39]. با انجام این کار، تعداد نقاط گوشه در محدودهٔ لب به تعداد $m'$ کاهش پیدا می‌کند؛ سپس دوباره بازشناسی حالت چهره از نمایش اشکال هندسی نقاط بالا توسط الگوریتم (۱) همانند روش قبلی انجام می‌گیرد.

$$Cov(x,y)=\frac{\sum_{i=1}^{n''}(x_i-\bar{x})(y_i-\bar{y})}{(n''-1)} \qquad (۱۳)$$

میانگین ارزش مقادیر پیکسل‌های یک تصویر با تعداد کل $n''$ پیکسل با $\bar{x}$ و $\bar{y}$ نشان داده شده است.

**(الگوریتم-۱): بازشناسی حالت چهره**

**(Algorithm-1): The facial expression recognition.**

**Input:** the number of extracted key-points in the mouth area: $n'$, and their two-dimensional coordinate values: $\mathbf{P}(x,y)$.

**Create** a vector $[1,n']$ in a Linear Space as $\mathbf{V}_{ls}$;
**Create** a Zero Matrix $[n',n']$ as $\mathbf{M}_z$;
**For** $(i=1; i \le n'; i++)$ **then** // consider each column of matrix $\mathbf{M}_z$
  **For** $(j=1; j \le n'; j++)$ **then** // consider each row of matrix $\mathbf{M}_z$
    **Fill** each column with: $\left(\frac{n'!}{(n'-j)!\,j!}\right)\circ(\mathbf{V}_{ls} \square j)\circ(1-\mathbf{V}_{ls}) \square (n'-j+1)$;
  **End For**
**End For**
**Create** a Modified Matrix $[n',1]$ as $\mathbf{M}_m = \mathbf{V}_{ls}^T \circ n'$;
**Compute** and **Permute** new two-dimensional coordinate values of the key-points: $\boldsymbol{P}'(x,y)=(\mathbf{M}_z+\mathbf{M}_m)\times \boldsymbol{P}(x,y)$;
**End For**
**If** $(\boldsymbol{P}'(x,y) > \boldsymbol{P}(x,y))$ **then**
  $\mathbf{F}=1$; // the facial expression is recognized
**Else**
  $\mathbf{F}=0$; // the facial expression is not recognized
**End if**

**Output:** return $\mathbf{F}$.

### ۳-۲-۳- سناریوی سوم: ترکیب Harris و BRISK

ایجاد یک ماژول ترکیبی از الگوریتم Harris و توصیف‌گر BRISK در این روش پیشنهاد شده است. بدین ترتیب که هر دو الگوریتم در دو فازکاری موازی به‌صورت جداگانه نقاط بحرانی محدودهٔ لب را استخراج کرده و در یک ترکیب مقایسه‌ای بازشناسی حالت لبخند و خنده چهره صورت می‌پذیرد. مرحلهٔ نخست همانند روش پیشنهادی یک، نقاط بحرانی مستخرج از محدودهٔ لب همانند رابطه (۱۲) ذخیره می‌شوند. در مرحلهٔ دوم، توصیف و استخراج نقاط بحرانی محدودهٔ لب توسط الگوریتم توصیف‌گر BRISK انجام می‌شود [40].

این الگوریتم توصیف‌گر در سه گام اصلی کار می‌کند، در گام نخست، ابتدا محدودهٔ فضای نقاط کلیدی بر پایه شدت روشنایی پیکسل‌ها توسط روش گوشه‌یاب تطبیق و آزمون بخش شتاب‌یافته عام و تطبیقی [1](AGAST) آشکار و سپس پیکسل‌ها در همسایگی نقطه مرکزی در قالب ناحیه فضاهای مقیاس[2] آشکار می‌شود؛ بدین ترتیب تصویر ورودی به تعدادی نمونه به‌نام اکتاوها و اکتاوهای درونی، بر پایهٔ الگوریتم (۲) با ورودی تصویر رابطه (۶) و بیشینهٔ مقدار $i$ برابر چهار ایجاد می‌شوند. در مرحلهٔ دوم برای هر کدام از اکتاوها و اکتاوهای درونی، نقاط کلیدی بر پایه روش گوشه‌یاب FAST استخراج می‌شوند. همان‌گونه‌که در شکل (۵) نشان داده شده‌است، در این روش، میزان شدت روشنایی، حول یک پیکسل مرکزی با پیکسل‌ها در همسایگی درجهٔ سوم آن مورد بررسی قرار می‌گیرد؛ درصورتی‌که میزان اختلاف شدت روشنایی بیش‌تر پیکسل‌های متوالی در این همسایگی با پیکسل مرکزی فاحش باشد، به‌عنوان یک نقطه کلیدی در نظر گرفته می‌شود.

[1] Adaptive and Generic Accelerated Segment Test
[2] Scale-space Area

**(الگوریتم-۲): آشکارساز ناحیه فضاهای مقیاس**

**(Algorithm-2): The facial expression recognition.**

**Input:** an original image that called octave $c_0$ .

**For** $(i=1; i \le n-1; i++)$ **then**

**Create** octave images, where octave $c_i$ is half-sampled from prior octave and compute scale: $t(c_i)=2^i$ ;

**Create** intra-octave images, where octave $d_i$ is down-sampled located between octaves $c_i$ and $c_{i+1}$ and compute scale: $t(d_i)=2^i \cdot 1.5$ ;

**End For**

**Output:** octave and intra-octave images.

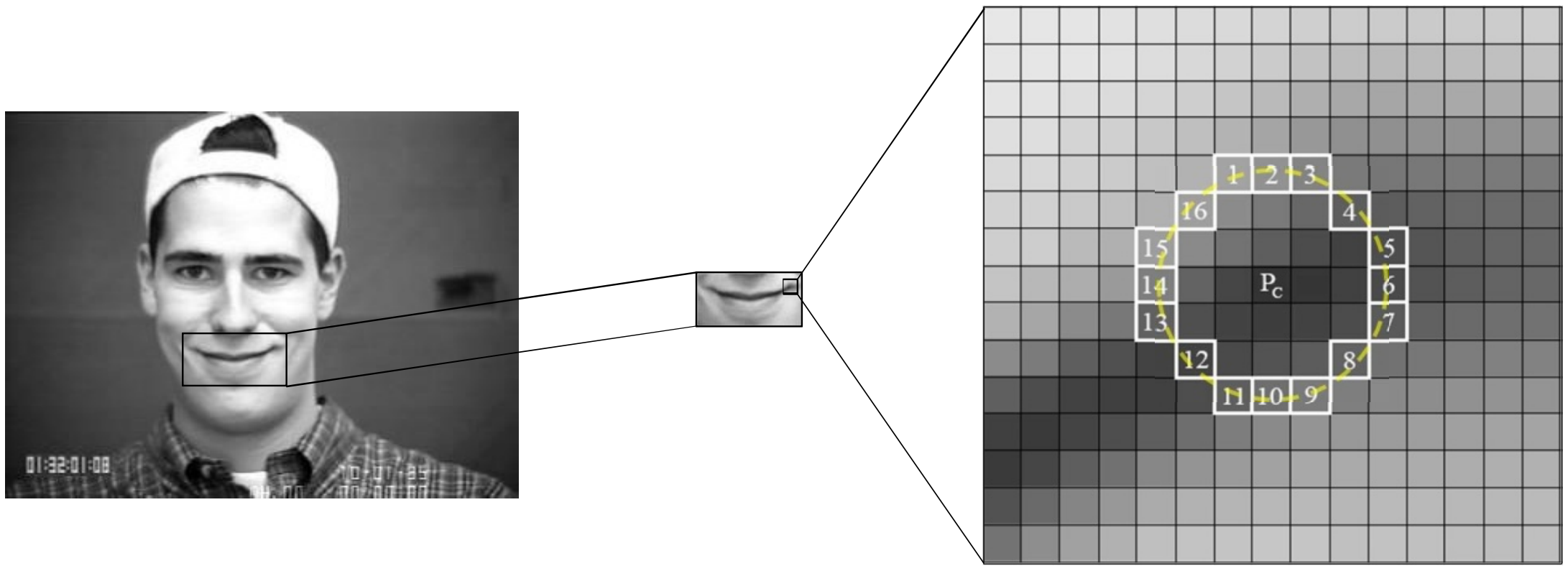

**(شکل-۵): شمای کلی از بررسی پیکسل‌های همسایهٔ درجهٔ سوم در روش گوشه‌یاب FAST، تصویر اصلی (سمت چپ)، محدوده لب آشکار شده (وسط) و پیکسل‌های گوشه سمت راست دهان (سمت راست)**

**(Figure-5): Overview of third adjacent pixel in FAST corner detector method: original image (Left), detected mouth area (middle), the pixels location on right corner of a mouth (right).**

توصیف نقاط کلیدی استخراج‌شده در مرحلهٔ سوم انجام می‌شود. در توصیف‌گر نقاط کلیدی[1]، ترکیب‌های دوتایی $(p_i,p_j)$ از این نقاط (رابطه (۱۴)) به‌صورت تصادفی انتخاب و بر اساس فاصله زوج نقاط انتخاب‌شده، به دو گروه نقاط با فاصلهٔ نزدیک و نقاط با فاصلهٔ دور که به‌ترتیب در رابطه‌های (۱۵) و (۱۶) نشان داده شده‌اند، تقسیم می‌شوند. با محاسبه گرادیان محلی مجموع زوج نقاط با فاصله دور از رابطه (۱۷)، جهت‌گرایی تصویر را از رابطه (۱۸) می‌توان محاسبه کرد و سپس به‌کمک رابطه (۱۹) مجموع زوج نقاط با فاصله نزدیک را بر اساس آن چرخاند؛ با این عمل یک توصیف‌گر دودویی ایجاد می‌شود. جهت محاسبات میزان اختلاف بین دو نقطه کلیدی در قسمت توصیف‌گر همخوان‌ساز[2] از تابع بولی XOR جهت کد فاصله همینگ آن‌ها استفاده می‌شود. اگر دو نقطه با یکدیگر شکل یکسان داشته باشند، در خروجی مقدار صفر و در غیر این‌صورت مقدار یک ظاهر می‌شود؛ بدین صورت نقاط بحرانی محدودهٔ لب استخراج می‌شوند.

$$A=\{(p_j,p_i)\in \mathrm{R}^2\times \mathrm{R}^2 \mid i,j\in N,\ j<i<N\} \qquad (۱۴)$$

$$S=\{(p_i,p_j)\in A \mid \|p_j-p_i\|<\sigma_{max}\}\subseteq A \qquad (۱۵)$$

$$L=\{(p_i,p_j)\in A \mid \|p_j-p_i\|>\sigma_{min}\}\subseteq A \qquad (۱۶)$$

$$G=\begin{pmatrix} g_x \\ g_y \end{pmatrix}=\frac{1}{L}\cdot \sum_{(p_i,p_j)\in L} G(p_i,p_j) \qquad (۱۷)$$

$$(p_j,p_i)_{Orientation}=G(p_j-p_i)\frac{I(p_j,\sigma_i)-I(p_j,\sigma_j)}{\|p_j-p_i\|^2} \qquad (۱۸)$$

$$L=\{(p_i,p_j)\in A \mid \|p_j-p_i\|>\sigma_{min}\}\subseteq A \qquad (۱۹)$$

ماژول بازشناسی حالات چهره بر پایه دو تابع برازش درجه دو ایجاد شده از مختصات نقاط کلیدی محدوده لب در الگوریتم‌های Harris و BRISK، و همچنین فاصله اقلیدسی

[1] Key-point Descriptor

[2] Descriptor Matching

میان نقاط بیشینه و کمینه این دو تابع می‌باشد. ابتدا توسط الگوریتم (۳)، مقادیر ضرایب ثابت و عرض از مبدا هر یک از این توابع به‌دست می‌آید؛ بسته به این مقادیر می‌توان طول نقطه کمینه و یا بیشینه هر تابع را توسط رابطه (۲۰) محاسبه نمود. لازم به ذکر است اگر $A_i > 0$ باشد آنگاه تابع کمینه دارد و بیشینه ندارد. و در صورتی که $A_i < 0$ باشد در این صورت تابع بیشینه دارد و کمینه نخواهد داشت و در نتیجه راس سهمی توابع به‌عنوان یک ویژگی قابل محاسبه است. دو ویژگی مهم در مورد سهمی توابع این است که مرکز تقارن ندارند و هر سهمی یا یک بیشینه دارد و یا یک کمینه، در نتیجه در پایان این مرحله مختصات دو نقطه محاسبه می‌گردد.

**(الگوریتم-۳): محاسبه پارامترهای بازشناسی حالت چهره در سناریو سوم**
**(Algorithm-3): The calculator of facial expression recognition parameters in the third scenario.**

**1: Input:** the number of extracted key-points in the mouth area using Harris and BRISK algorithms: $n_1$ and $n_2$, and their two-dimensional coordinate values: $P_1(x,y)$ and $P_2(x,y)$ respectively.
**2:** **Find** the maxima and minima coordinate values of $P_2(x,y)$ as $Max_2(x,y)$ and $Min_2(x,y)$;
**3:** **Calculate** the average of the values: $Average_2(x,y) = (Max_2(x,y) + Min_2(x,y))/2$;
**4:** **Calculate** the new coordinate values: $P_2'(x,y) = Average_2(x,y) \times P_2(x,y)$;
**5:** **Interpolate** the coordinate values of $P_2'(x,y)$ by using Spline Interpolation method as $P_{2i}'(x,y)$;
**6:** **Find** the coefficients of a polynomial $P_2'(x,y)$ of degree 2 as $P_2'(x,y) = A_2x^2 + B_2x + C_2$;
**7:** **Do** step 6 for Harris coordinate values: $P_1(x,y) = A_1x^2 + B_1x + C_1$;
**8: Output: For** $(i=1; i \le 2; i++)$ **then** return $A_i, B_i, C_i$ **End For.**

$$\begin{cases} \Delta_i = B_i^2 - 4A_iC_i \\ P_i(x,y) = \begin{vmatrix} x_i = -B_i/2A_i \\ y_i = -\Delta_i/4C_i \end{vmatrix} \end{cases} \quad where\ i\ is\{1,2\} \qquad (۲۰)$$

پس از محاسبه مختصات نقاط بیشینه و کمینه توابع ذکر شده، فاصله اقلیدسی بین آن‌ها بر اساس رابطه (۲۱) محاسبه می‌شود. میزان طول فاصله این نقاط از همدیگر بیان‌گر حالات چهره است.

$$Dist(x,y) = \sqrt{\sum_{i=1}^{2}(x_i - y_i)^2} \qquad (۲۱)$$

## ۴- نتایج پیاده‌سازی و ارزیابی

همان‌طورکه در بخش‌های پیشین توضیح داده شد، در این پژوهش تصاویر سه حالت احساسی خنده، لبخند و خنثی از پایگاه‌داده Cohn-Kaonde جهت بررسی کارایی روش پیشنهادی مورد استفاده قرار گرفته است. همچنین برای پیاده‌سازی و شبیه‌سازی روش ارائه‌شده، از نرم‌افزار MATLAB R2015b بر روی سامانه‌ای با پردازش‌گر Core i5 5200U 2.4 GHz، و حافظه 4GB بهره گرفته شده است.

ابتدا جهت کاهش پیچیدگی محاسباتی و همچنین ازدست‌ندادن جزییات بافت تصاویر، هر تصویر ورودی به بیشینه سطح خاکستری یعنی ۲۵۶ سطح تغییر داده و سپس جهت تقویت تفاوت پیکسل‌ها و تمایز بین ویژگی آن‌ها الگوریتم E-AIHT اعمال می‌شود. در این الگوریتم، پارمترهای $\alpha$ و $\beta$ تابع کنترلی Bias مطابق شکل (۶) به‌ترتیب ۰/۱۲۵ و ۰/۲۵ در نظر گرفته می‌شوند؛ درنتیجه میزان متوسط سرعت تغییرات روشنایی پیکسل‌ها جهت تبدیل تصاویر به سطح خاکستری مطلوب به‌دست می‌آید. تابع وزنی Gain درواقع میزان پراکندگی شدت روشنایی پیکسل‌ها را در تصاویر گرداگرد مقدار میانگین آن‌ها تعدیل می‌کند. همان‌طور که شکل (۷) نشان می‌دهد، جهت محاسبهٔ متوسط پراکندگی شدت روشنایی، پارمترهای $\rho$ و $\gamma$ به‌ترتیب ۰/۱ و ۰/۵ می‌تواند در نظر گرفته شود. در انتهای این مرحله، تصویر ورودی به تصویری مطلوب با سطح خاکستری تبدیل می‌شود که در کارایی مؤثر مراحل بعدی سامانه نقش مستقیمی می‌تواند داشته باشد. شکل (۸) نمونه‌های از داده‌های ورودی و خروجی این مرحله را نشان می‌دهد.

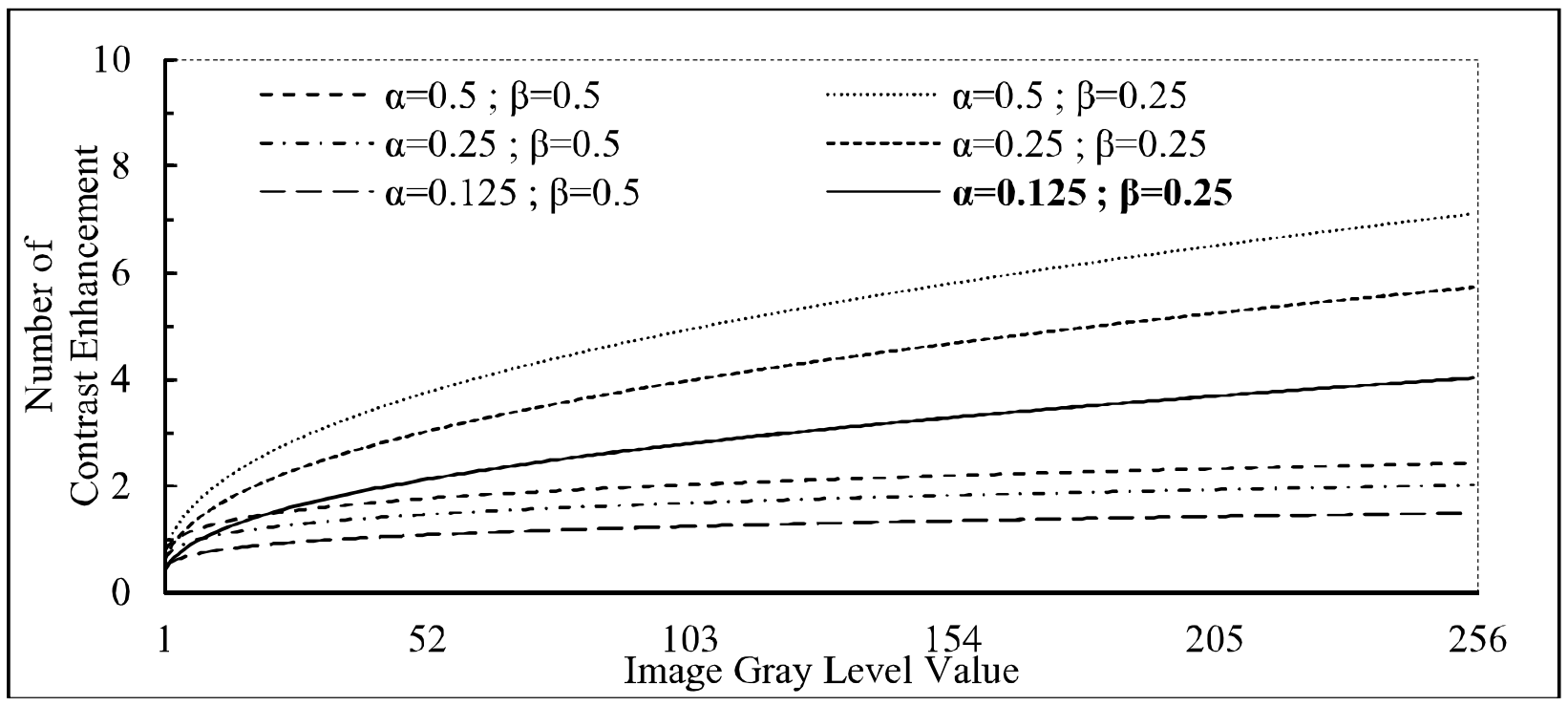

**(شکل-۶ ): نتایج تنظیم پارامترهای تابع کنترلی Bias**

**(Figure-6): The adjustment parameters results for bias control function.**

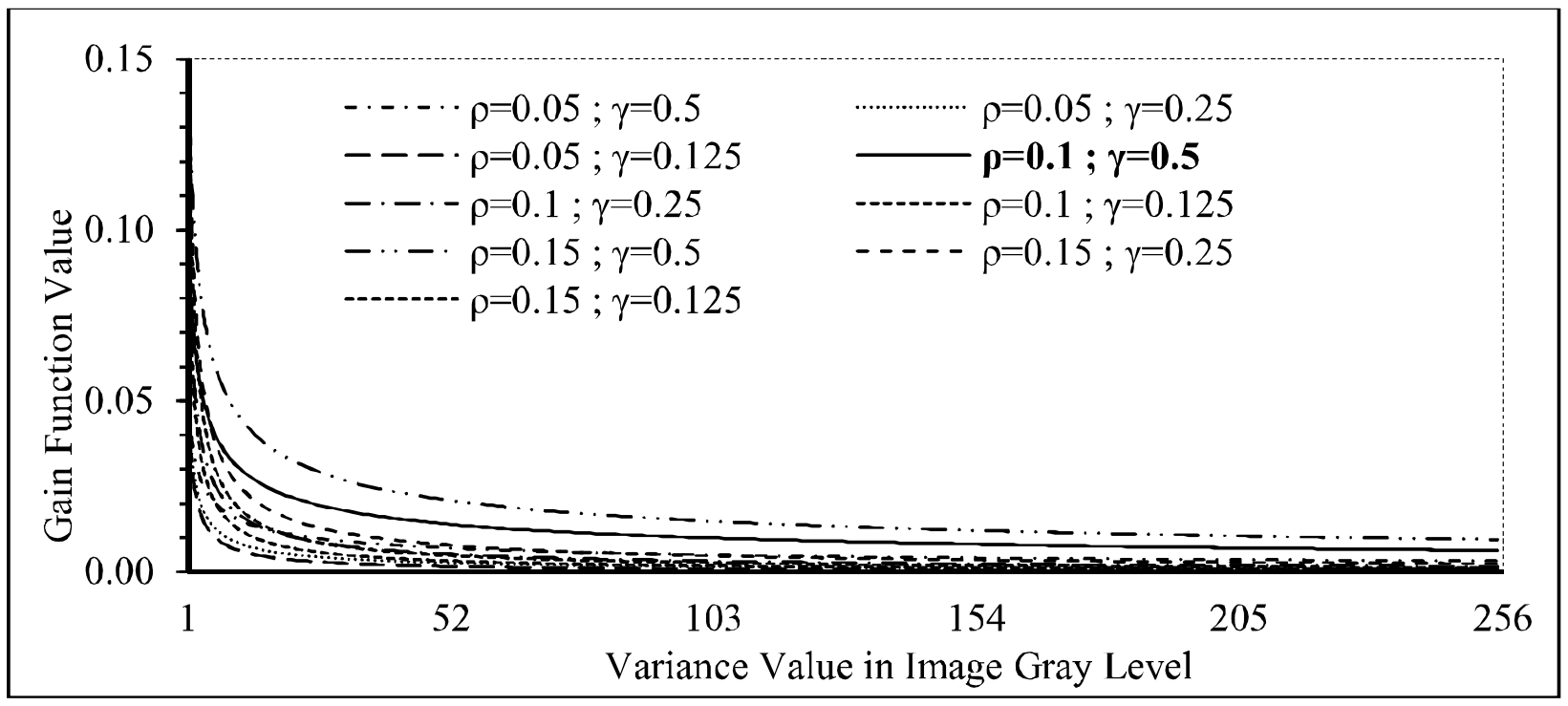

**(شکل-۷): نتایج تنظیم پارامترهای تابع وزنی Gain**

**(Figure-7): The adjustment parameters results for gain control function.**

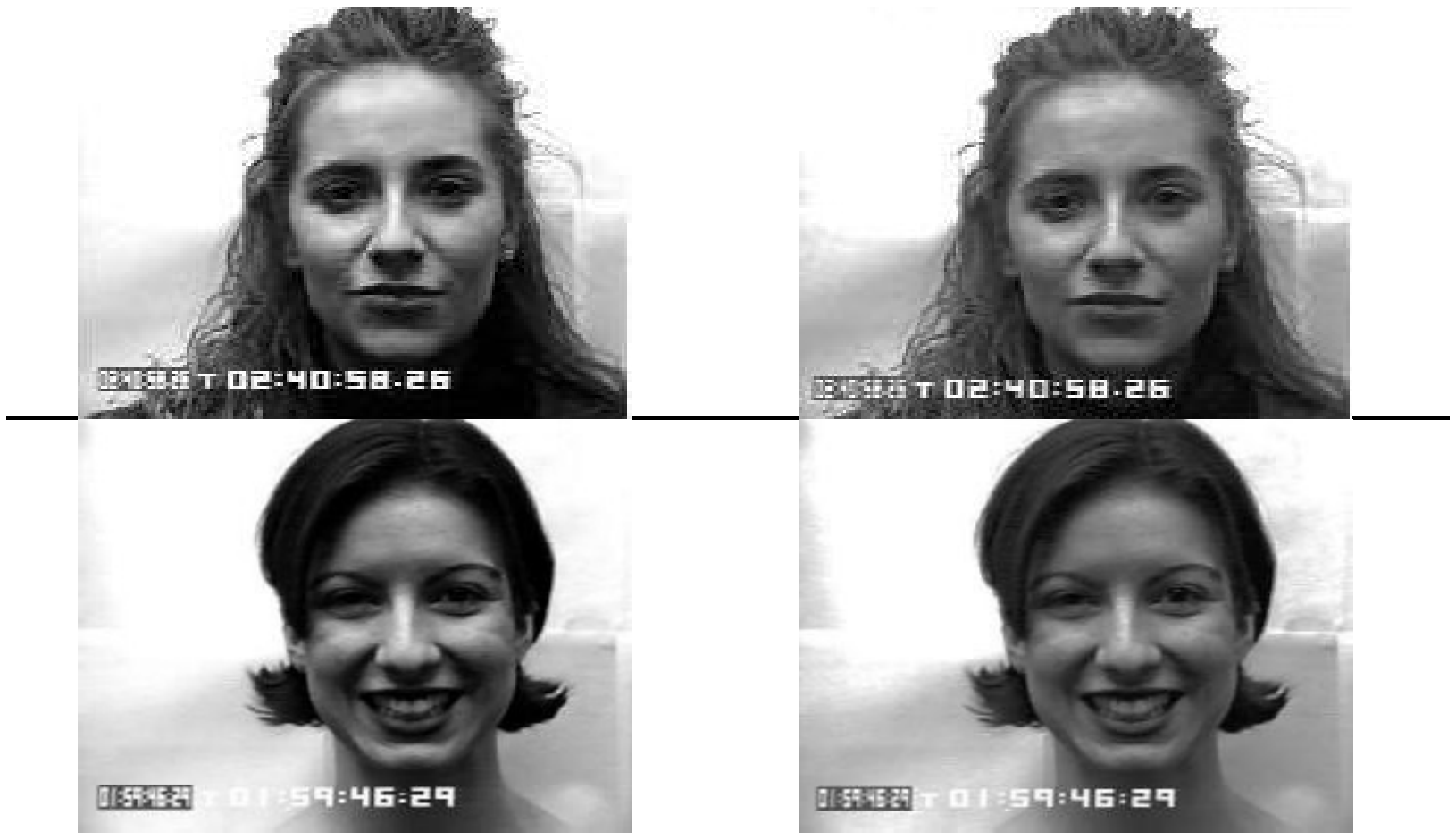

**(الف)** **(ب)**

**(شکل-۸): نتایج مرحله پیش پردازش: الف) تصاویر ورودی خام، ب) تصاویر خروجی**

**(Figure-8): The pre-processing step results: a) raw input images, b) output images.**

در گام بعدی، تشخیص حالات لبخند و خندهٔ چهره به‌کمک ویژگی‌های محدود لب انجام می‌پذیرد؛ درنتیجه در این مرحله جهت کاهش پیچیدگی پردازش با روش تحلیل بافت LBP این محدوده آشکار می‌شود. در این روش، تصویر بهبودداده‌شده، ابتدا به زیرتصاویری با پنجره‌های ۲۴×۲۴ پیکسل تقسیم‌بندی شده و سپس ویژگی‌های خط، لبه و

مستطیل شکل در این پنجره‌ها استخراج می‌شوند. ارزیابی ویژگی‌ها توسط تصاویر انتگرال با عمق یک انجام و انتخاب بهترین ویژگی‌ها جهت آشکارسازی محدودهٔ لب با الگوریتم یادگیری آدابوست انجام می‌شود [34,35]. شکل (۹) خروجی این مرحله را نشان می‌دهد.

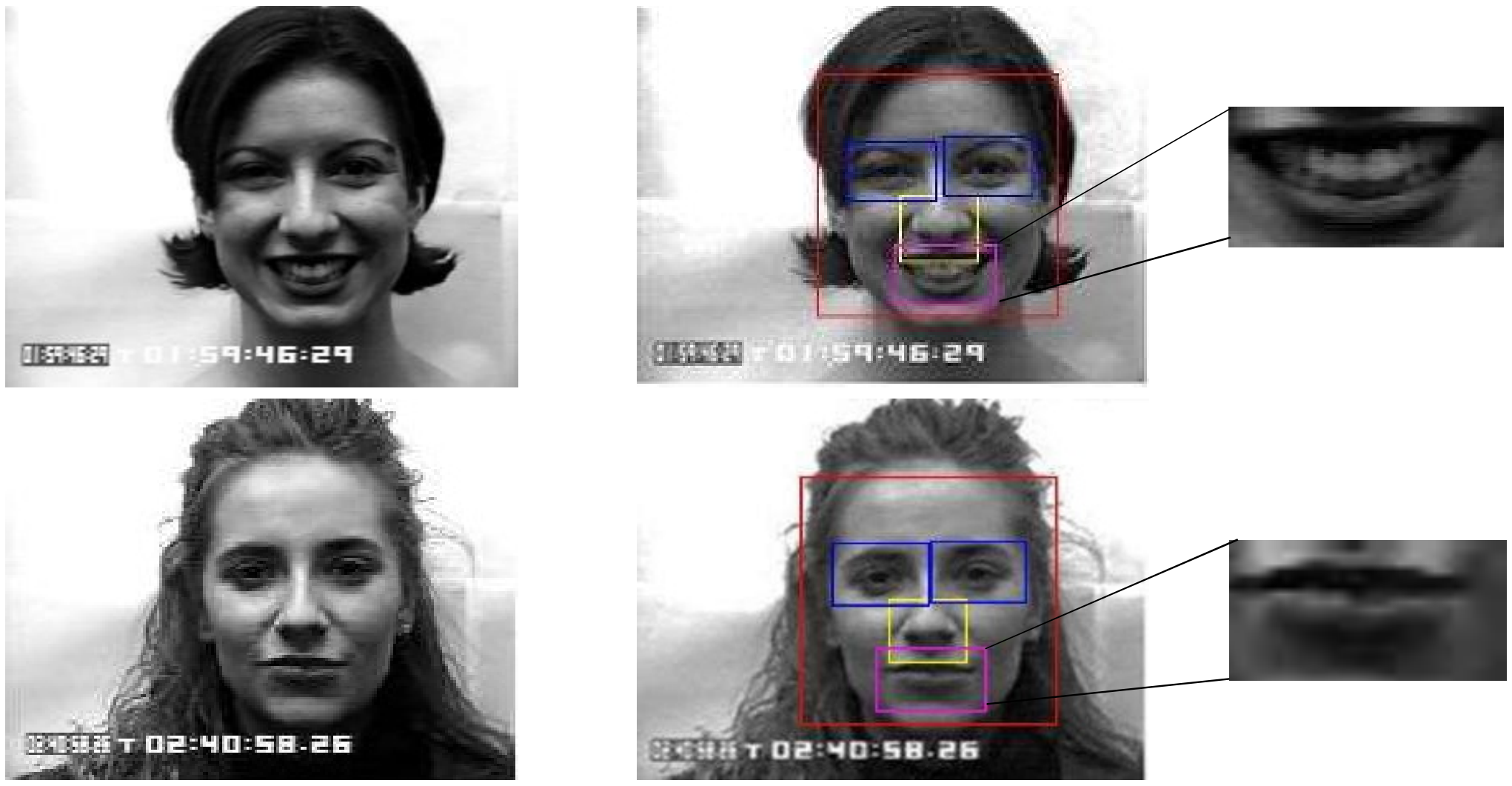

**(شکل-۹): نتایج آشکارسازی محدوده لب: الف) تصویر ورودی، ب) آشکارسازی اجزا چهره، ج) آشکارسازی محدوده لب**

**(Figure-9): The mouth area detection results: a) input image, b) facial parts detection c) mouth area detection.**

سه سناریوی مختلف جهت استخراج نقاط بحرانی محدودهٔ لب آشکارسازی‌شده بر پایهٔ الگوریتم Harris ارائه شده است و سپس بازشناسی حالات خنده و لبخند برای هر یک از سناریوها پیشنهاد شد. در سناریوی نخست، الگوریتم Harris گوشه‌های هندسی تصویر را از گوشه‌های بافت یا نوفه در محدودهٔ لب متمایز کرده و به‌عنوان نقاط بحرانی و یا ویژگی‌های محلی (الگوریتم (۱)) بر می‌گرداند. برای پیاده‌سازی الگوریتم Harris، ابتدا اندازه پنجره گوسی رابطه (۷) با مقدار $\sigma$ برابر ۱/۵ به‌دست آمده و سپس گرادیان یا میزان تغییر جهت در هر پنجره تصویر بر روی محور $x$ و $y$ در رابطه (۸) محاسبه می‌شوند. در ادامه، مقدار پارامتر $k$ در تابع پاسخ (رابطه (۹)) نیز در مطلوب‌ترین حالت برابر ۰/۰۴ در نظر گرفته شده و در صورت بزرگ‌بودن مقدار تابع پاسخ با حد آستانه پنجاه‌هزار، مشخص می‌شود یک پیکسل در ناحیه گوشه قرار دارد (شکل (۱۰))؛ در مرحلهٔ بازشناسی این سناریو، نقاط گوشهٔ استخراج‌شده به‌عنوان نقاط کنترلی منحنی بزیر در نظر گرفته می‌شود؛ در واقع خاصیت پوسته محدب یک ویژگی مهم منحنی بزیر است که در ادامه توسط تابع برنشتاین تفسیر شکل نقاط به‌دست آمدهٔ منحنی که همان بازشناسی حالت چهره است، انجام می‌گیرد. در سناریوی دوم، بر روی نقاط استخراج‌شده از الگوریتم (۱)، PCA را اعمال کرده تا در یک فضای برداری با کاهش مجموعه نقاط گوشه آشکار شوند، نقاطی که بیشترین تأثیر در واریانس را دارند، حفظ شده و مابقی حذف شوند؛ مرحله بازشناسی در این مرحله نیز همانند سناریوی نخست انجام شد. در سناریوی سوم، ماژول ترکیبی الگوریتم Harris و الگوریتم BRISK در دو فاز کاری موازی به صورت جداگانه وظیفه آشکارسازی نقاط بحرانی محدوده لب تصاویر را بر عهده دارند. در مرحلهٔ نخست همانند سناریوی نخست نقاط بحرانی استخراج می‌شود. در مرحلهٔ دوم، توصیف نقاط بحرانی محدودهٔ لب در الگوریتم BRISK تا عمق چهار نمونه اکتاوها بر پایه الگوریتم (۲) انجام می‌شود؛ بدین صورت که میزان شدت روشنایی حول یک پیکسل مرکزی با پیکسل‌ها در همسایگی درجه سوم هر نمونه اکتاو مقایسه می‌شود. درصورتی‌که از حد آستانه ۰/۰۱ بیشتر باشد، به‌عنوان پیکسل کلیدی در نظر گرفته می‌شود. با محاسبه مجموعه فاصله نزدیک ($\sigma_{min}$ < ۳۲بیت) و فاصله دور (۱۲۸بیت < $\sigma_{max}$) پیکسل‌های کلیدی به دو گروه تقسیم می‌شوند. با محاسبه گرادیان محلی مجموع زوج نقاط با فاصله دور جهت‌گرایی تصویر را می‌توان توصیف و همچنین با محاسبه فاصله همینگ بین نقاط نامزد کلیدی، مجموعه نقاط بحرانی را توصیف کرد. در مرحله بازشناسی این سناریو، ابتدا تابع برازش نقاط کلیدی استخراج‌شده از الگوریتم‌های Harris و

BRISK در قالب دو معادله درجهٔ دو به‌دست آمد؛ سپس نقاط بیشینه و کمینه هر معادله محاسبه شد. بر پایه توصیف رقمی فاصله اقلیدسی این نقاط و به‌کمک اعداد تجربی به‌دست‌آمده در جدول (۲)، حالات چهره بازشناسی شد. به‌عنوان مثال، همان‌گونه‌که در سطر نخست جدول (۲) مشاهده می‌شود، درصورتی‌که هر دو مقادیر منفی ($A_i < 0$) و فاصله اقلیدسی دو تابع درجه دو سهمی بزرگ‌تر از ۲۵۰۰ باشد، آن‌گاه حالت چهره بازشناسی می‌شود. درضمن اگر مقادیر $y_i$ها خیلی بزرگ باشند، بازشناسی چهره اتفاق نخواهد افتاد. نتایج مقدار کمینه و بیشینهٔ نقاط بحرانی در محدودهٔ لب تصاویر پایگاه داده‌ها در جدول (۳) نشان داده شده است. در جدول (۴) مقادیر به‌دست‌آمده برای میزان حساسیت سامانه نسبت به چرخش تصاویر نشان داده شده است؛ نتایج شبیه‌سازی، بیان‌گر بهبود مقاومت سامانه در برابر چرخش تصاویر با وجود کاهش بعد استخراج نقاط کلیدی است، درواقع این بدان دلیل است که شکل هندسی و فاصله نقاط کلیدی از هم‌دیگر در بازشناسی سامانه نقش اصلی را ایفا می‌کنند. ذکر این نکته ضروری به نظر می‌رسد که اگر چه سناریوهای پیشنهادی در راستای افزایش دامنه چرخش تصاویر و کاهش پیچیدگی محاسباتی است، اما فاصله تا چرخش کامل از معایب آن است.

**(جدول-۲): میزان طول فاصله اقلیدسی نقاط بیشینه و کمینه توابع درجه دوم سهمی**

**(Table-2): Euclidean distance from the minimum and maximum points of quadratic equations.**

| حالت | $A_2 < 0$ | $A_1 < 0$ | $Dist(x,y)$ |
|---|---|---|---|
| اول | 0 | 0 | >2500 |
| دوم | 1 | 0 | >3000 |
| سوم | 0 | 1 | >2000 |
| چهارم | 1 | 1 | 7000< یا 5000> |

نرخ پراکندگی خطای سامانه در جدول (۵) نشان داده شده است؛ به‌طور تقریبی نیمی از خطاهای سامانه مربوط به تصاویری است که در موقعیت و شرایط نامطلوب همچون نمای دور چهره و یا اختلاف شدت نور زیاد تصویربرداری و تهیه شده‌اند؛ و یا حدود ۲۰٪ تصاویری که چهره‌های آن‌ها هم‌زمان دارای ریش و سبیل بوده‌اند، به‌دلیل مشخص‌نبودن نقاط کلیدی گوشه‌های لب به‌اشتباه تشخیص داده شده‌اند؛ از طرفی دیگر، خطاهای باقیمانده سامانه را می‌توان به‌طورتقریبی به تساوی ناشی از وجود چین و چروک و یا فقط سبیل در چهره‌ها گزارش کرد.

**(جدول-۳): تعداد کمینه و بیشینه نقاط بحرانی در تصاویر پایگاه داده‌ها**

**(Table-3): The minimum and maximum values of critical points in the different image datasets.**

| روش | تعداد نقاط کلیدی مستخرج در پایگاه‌داده | | | | | | | |
|---|---|---|---|---|---|---|---|---|
| | Cohn-Kanade | | JAFFE | | Yale | | CAFÉ | |
| | کمینه | بیشینه | کمینه | بیشینه | کمینه | بیشینه | کمینه | بیشینه |
| **سناریوی نخست:Harris** | 1 | 168 | 10 | 204 | 8 | 519 | 10 | 640 |
| **سناریوی دوم: PCA + Harris** | 1 | 113 | 8 | 45 | 7 | 90 | 8 | 96 |
| **سناریوی سوم: BRISK + Harris** | 1 | 20 | 7 | 40 | 7 | 81 | 6 | 90 |

**(جدول-۴): میزان کمینه و بیشینه چرخش در تصاویر پایگاه داده‌ها**

**(Table-4): The maximum and minimum values of rotation in the different image datasets.**

| روش | سامانهٔ مقاوم نسبت به چرخش در پایگاه‌داده | | | | | | | | | |
|---|---|---|---|---|---|---|---|---|---|---|
| | Cohn-Kanade | | JAFFE | | Yale | | CAFÉ | | میانگین کمینه | میانگین بیشینه |
| | کمینه | بیشینه | کمینه | بیشینه | کمینه | بیشینه | کمینه | بیشینه | | |
| **سناریوی نخست:Harris** | 2 | 40 | 3 | 30 | 5 | 48 | 3 | 46 | 3.25 | 41 |
| **سناریوی دوم: PCA + Harris** | 7 | 53 | 6 | 47 | 7 | 56 | 7 | 54 | 6.75 | 52.5 |
| **سناریوی سوم: BRISK + Harris** | 8 | 62 | 10 | 48 | 8 | 55 | 13 | 68 | 9.75 | 58.25 |

(جدول-۵): نرخ خطای تشخیص در تصاویر پایگاه داده‌ها

(Table-5): Recognition error rate in the different image datasets.

| نوع خطا | سناریو اول: Harris | | | | سناریو دوم: PCA + Harris | | | | سناریو سوم: BRISK + Harris | | | |
|---|---|---|---|---|---|---|---|---|---|---|---|---|
| | *Cohn Kaonde* | *JAFFE* | *Yale* | *CAFÉ* | *Cohn Kaonde* | *JAFFE* | *Yale* | *CAFÉ* | *Cohn Kaonde* | *JAFFE* | *Yale* | *CAFÉ* |
| سبیل | 0.113 | --- | 0.014 | 0.017 | 0.083 | --- | 0.013 | 0.007 | 0.018 | --- | 0.009 | 0.005 |
| ریش و سبیل | 0.057 | --- | 0.041 | 0.052 | 0.11 | --- | 0.032 | 0.029 | 0.028 | --- | 0.014 | 0.015 |
| صورت مسن، چین و چروک | 0.028 | 0.068 | 0.027 | 0.035 | 0.028 | 0.018 | 0.006 | 0.021 | 0.009 | 0.028 | 0.005 | 0.014 |
| تصاویر نامطلوب | 0.142 | 0.113 | 0.068 | 0.086 | 0.11 | 0.123 | 0.019 | 0.043 | 0.055 | 0.083 | 0.023 | 0.036 |

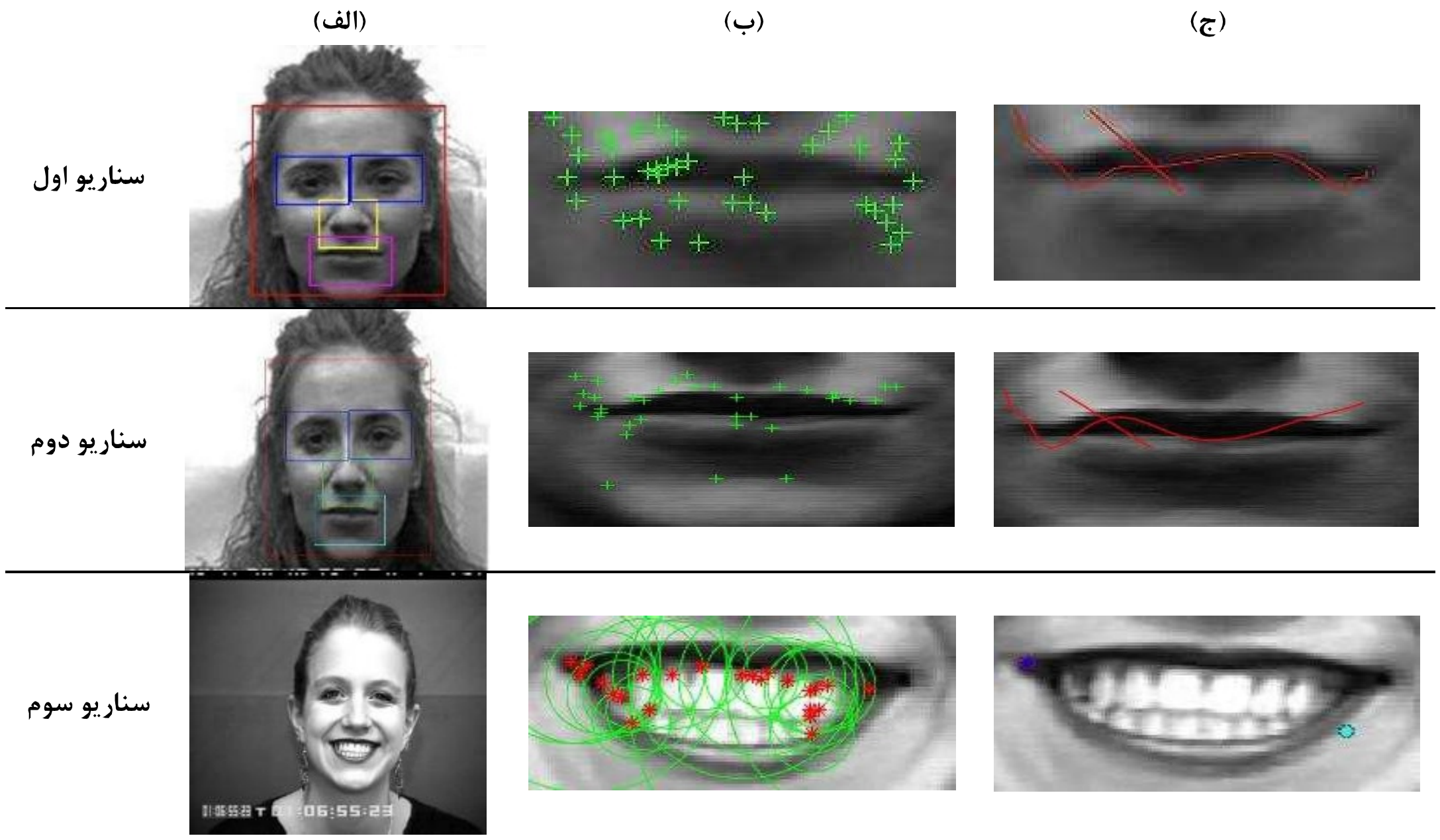

(شکل-۱۰): کمینه و بیشینه نقاط گوشه استخراج شده در محدوده لب: الف) آشکارسازی اجزا چهره، ب) استخراج نقاط کلیدی محلی، ج) آشکارسازی حالت چهره

(Figure-10): The minimum and maximum number of the extracted corner points in the mouth area: a) facial parts detection, b) extracting local key points, c) facial expression detection.

در این قسمت به ارزیابی سه سناریوی مختلفی که سعی کرده‌اند، نقاط بحرانی را بر پایه گوشه‌یابی، کاهش ابعاد داده و توصیف‌گر نقاط متمایز کنند، پرداخته می‌شود. نتیجهٔ بازشناسی حالات لبخند و خنده چهره این سناریوها بر روی دادگان پژوهش به‌ترتیب در شکل (۱۱) نشان داده شده است. این نمودارها بر اساس دو معیار فراخوانی[1] و دقت[2] با اجرای تصادفی بر روی تصاویر پایگاه‌داده با تنظیمات ذکرشده در قسمت پیادسازی سامانه گزارش شده‌اند. این نتایج نشان می‌دهد که استخراج زیاد تعداد نقاط بحرانی منجر به افزایش نرخ دقت بازشناسی سامانه نمی‌شود و فقط نقاط بحرانی حاوی اطلاعات معنادار نقش مستقیمی در بالابردن کارایی سامانه می‌توانند داشته باشند. جدول (۶) نرخ متوسط بازشناسی حالات خنده و لبخند بر روی دادگان پژوهش را نشان می‌دهد که نتایج تأیید می‌کنند با هدایت کلی سامانه به استخراج نقاط با مؤلفه‌های ذاتی معنادار هم پیچیدگی محاسبات را کاست و هم نرخ کارایی سامانه را می‌توان افزایش داد و همچنین عملکرد کلی سامانه را قابل قبول ارزیابی می‌کند.

[1] Recall

[2] Precision

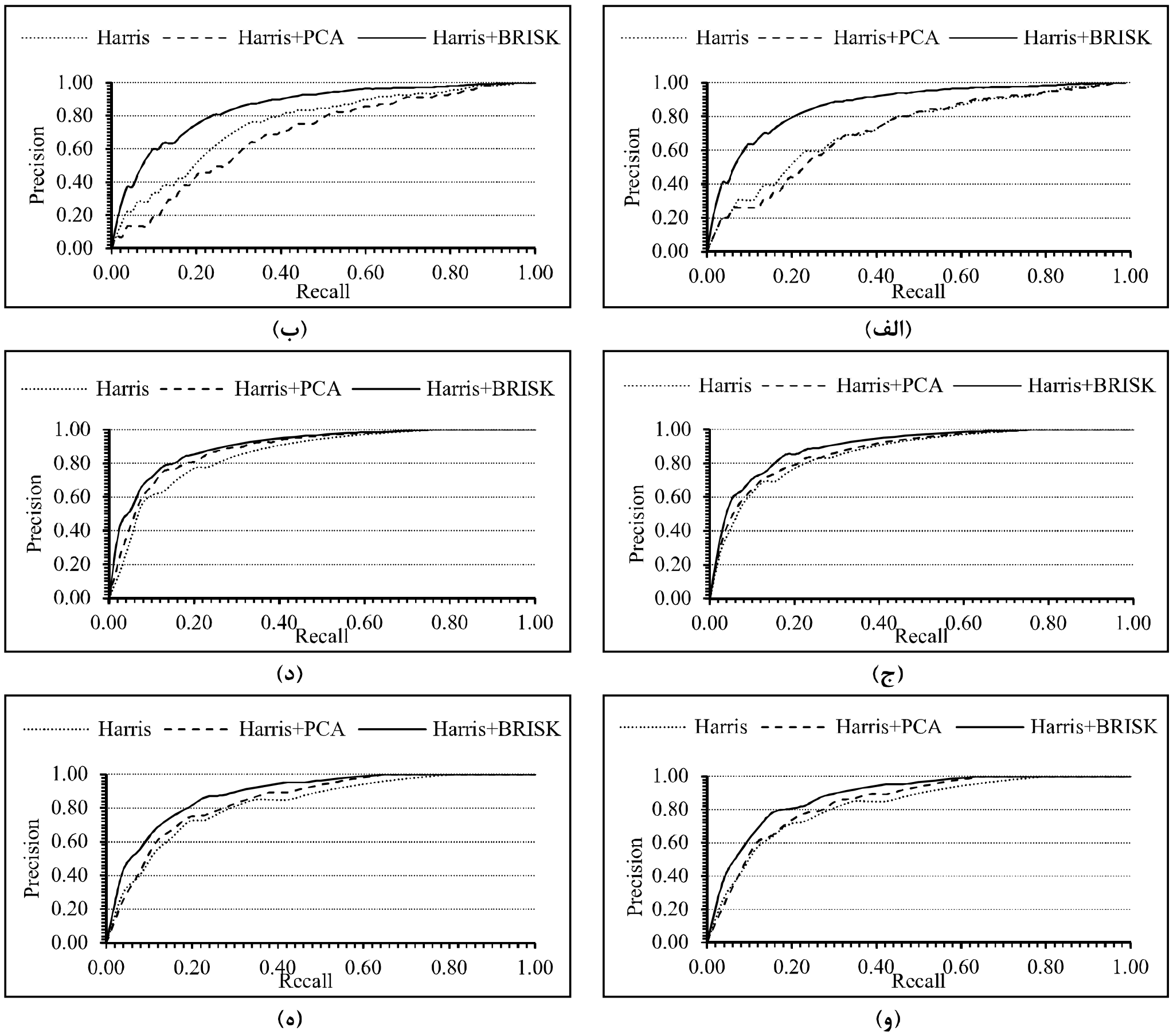

**(شکل-۱۱): نتایج بازشناسی حالات چهره در پایگاه داده‌های مختلف: الف) Cohn-Kaonde (لبخند)، ب) Cohn-Kaonde (خنده)، ج) CAFE (لبخند)، د) CAFE (خنده)، و) JAFFE ه) Yale**

**(Figure-11): The facial expression recognition results in the different databases: a) Cohn-Kaonde (smile), b) Cohn-Kaonde (laugh), j) CAFE (smile), d) CAFE (laugh), v) JAFFE, h) Yale**

**(جدول-۶): نتایج نرخ متوسط بازشناسی حالات چهره در پایگاه داده‌های مختلف و کل سیستم**

**(Table-6): The overall percentage of facial expression recognition in the different databases.**

| روش | متوسط نرخ دقت بازشناسی در حالت‌های خنده و لبخند (٪) | | | | |
|---|---|---|---|---|---|
| | Cohn-Kanade | JAFFE | CAFE | Yale | کل سیستم |
| سناریوی نخست: Harris | 0.66 | 0.82 | 0.85 | 0.81 | 0.79 |
| سناریوی دوم: Harris + PCA | 0.67 | 0.86 | 0.93 | 0.9 | 0.84 |
| سناریوی سوم: Harris + BRISK | 0.89 | 0.89 | 0.95 | 0.93 | 0.92 |

جدول (۷) مقایسه روش‌های مختلف را نشان می‌دهد. نتایج نشان می‌دهد با کاهش ابعاد استخراج ویژگی‌ها، پیچیدگی محاسبات نیز کم می‌شود و سرعت اجرا افزایش می‌یابد؛ همچنین از نقطه‌نظر ارزیابی کمی، بعضی از روش‌ها دارای نرخ دقت تشخیص بالاتر هستند؛ ولی از نظر کیفی میزان مقاومت در برابر چرخش و یا متوسط زمان اجرا گزارش نشده است و پایه روش پیشنهادی آن‌ها بر آموزش سامانه است؛ به‌گونه‌ای که تطبیق حالت چهره در آزمون با حالت چهره آموزش‌دیده وابستگی به خصوصیات ذاتی هر پایگاه داده تصاویر را نشان می‌دهد. ذکر این نکته ضروری به نظر می‌رسد که اگر تعداد نقاط مستخرج زیاد باشد، درنتیجه تعداد نقاط با ویژگی‌های نامفید افزایش می‌یابد و به‌طور طبیعی باعث

نامطلوب‌شدن زمان محاسبات می‌شوند؛ از طرفی دیگر ماهیت اصلی الگوریتم‌های توصیف‌گر و استخراج‌کننده ویژگی‌های تصاویر استفاده‌شده در این پژوهش، بدین صورت است که در یک محدوده خاص وضوح تصویر بهترین کارآیی را دارند، که این بسته به شیء هدف، نوع و بافت تصویر دارد؛ و هر چقدر تصاویر ورودی در این محدوده دارای وضوح بالاتری باشند دقت و تعداد استخراج نقاط کلیدی مطلوب‌تر خواهد بود. در پایان به‌طور میانگین در صورت تبدیل مقیاس تصاویر اصلی به اندازه کمینه ۰/۴ و بیشینهٔ پنج برابر آن، روش پیشنهادی به‌خوبی کار می‌کند. همان‌طورکه در این جدول مشاهده می‌شود، میزان میانگین مقاومت در برابر چرخش الگوریتم پیشنهادی در بهترین حالت ۳۴± است، این در حالی است که این ارزیابی مقاومت در الگوریتم‌های سایر پژوهش‌گران ارائه نشده است. همچنین، بالاترین نرخ دقت بازشناسی و کمترین نرخ زمان اجرای به‌دست‌آمده مربوط به سناریوی سوم است که در مقایسه با کارهای دیگران، به‌طور میانگین بر روی تمامی پایگاه داده‌ها بالاترین نرخ دقت بازشناسی ۹۱/۵ درصد و نرخ زمان اجرای ۰/۷۷۱ ثانیه بر تصویر را دارد؛ از طرف دیگر همان‌طور که در این مقایسه مشاهده می‌شود، نرخ دقت بازشناسی سناریوی پیشنهادی سوم در پایگاه‌داده‌های CAFE و Yale بیشتر از سایرین، در پایگاه داده‌های JAFFE و Cohn-Kaonde به‌ترتیب به‌طور تقریبی برابر و کمتر از نرخ‌های گزارش‌شده توسط [41] است. هر چند الگوریتم پیشنهادی در بهترین حالت توانسته در دو پایگاه‌داده نسبت به کارهای دیگران بهتر، در دوتایی دیگر نیز برابر و یا نرخ بازشناسی کمتری به‌دست آورد، اما درمجموع با توجه به زمان اجرای کمتر الگوریتم پیشنهادی برتری محسوسی را در مقایسه با پژوهش‌های دیگران دارد.

**(جدول-۷): مقایسه کمی نتایج روش پیشنهادی با کارهای مرتبط**

**(Table-7): Quantitative comparison results of the proposed method with related works.**

| منابع | پایگاه داده | میانگین مقاومت در برابر چرخش | بازشناسی | آموزش سیستم | دقت سیستم (٪) | مشخصات نرم افزار و سیستم | زمان اجرا (ثانیه بر تصویر) |
|---|---|---|---|---|---|---|---|
| [42] | JAFFE | --- | Enhanced LBP | بلی | 84 | --- | --- |
| [43] | JAFFE | --- | Gabor wavelet representation + LDA | بلی | 78 | --- | --- |
| [41] | JAFFE<br>Cohn-Kaonde<br>Yale | --- | PCA+ICA & LDA+HMM | بلی | 88.7<br>86.1<br>86.8 | SW: Matlab, CPU: Intel Pentium Dual-Core™ 2.5 GHz, RAM: 3GB | 1.533<br>2.292<br>1.498 |
| | | | Wavelet Transform + HMM | | 90<br>94<br>88 | | 1.29<br>1.908<br>1.034 |
| [44] | JAFFE<br>Cohn-Kaonde | --- | Gabor + PCA | بلی | 89.67<br>91.51 | SW: ---, CPU: Intel Core i3 2.67 GHz, RAM: 2G | --- |
| روش پیشنهادی (سناریوی نخست) | Cohn-Kaonde<br>JAFFE<br>CAFE<br>Yale | ±22.13 | Harris | خیر | 66<br>85<br>82<br>81 | SW: MatlabR2015b, CPU: Intel Core i5 2.4 GHz, RAM: 4GB | 2.751<br>2.194<br>7.233<br>7.089 |
| روش پیشنهادی (سناریوی دوم) | | ±29.63 | PCA + Harris | | 67<br>86<br>93<br>90 | | 2.446<br>2.023<br>3.376<br>2.675 |
| روش پیشنهادی (سناریوی سوم) | | ±34 | BRISK + Harris | | 89<br>89<br>95<br>93 | | 0.83<br>0.752<br>0.743<br>0.758 |

## ۵- نتیجه‌گیری و کارهای آینده

در این مقاله سامانه‌ای بر پایه سه سناریوی مختلف استخراج نقاط بحرانی محدودهٔ لب پیشنهاد شد. این سناریوها بر اساس روندی که موجب افزایش شانس استخراج نقاط بحرانی که حاوی اطلاعات معنادار هستند ارائه شدند. همان‌طورکه بیان شد، کمترین تعداد این نقاط استخراج‌شده و بیشترین نرخ دقت بازشناسی مربوط به سناریوی Harris+BRISK بود که علاوه‌بر حاصل‌شدن نتایج قابل قبول، تأثیر به‌کارگیری

الگوریتم‌های توصیف‌گر در کاهش تعداد نقاط مستخرج و بالا بردن نرخ دقت بازشناسی سامانه را نشان می‌دهد. جهت ارزیابی کلی سناریوهای سامانهٔ پیشنهادی، به‌طور میانگین بالاترین نرخ دقت بازشناسی ۹۱/۵ درصد و نرخ زمان اجرای ۰/۷۷۱ ثانیه بر تصویر بر روی کلیه تصاویر پایگاه داده‌های شاخص به‌دست آمد. در پایان، به‌کارگیری دیگر الگوریتم‌های توصیف‌گر در سامانهٔ بازشناسی حالات مختلف چهره به‌عنوان پژوهش‌های آینده مورد بررسی می‌تواند قرار گیرد. همچنین جهت ارتقای سامانهٔ پیشنهادشده از الگوریتم‌های یادگیری عمیق جهت طبقه‌کننده نقاط بحرانی می‌توان استفاده کرد.

## 6- References ۶- مراجع

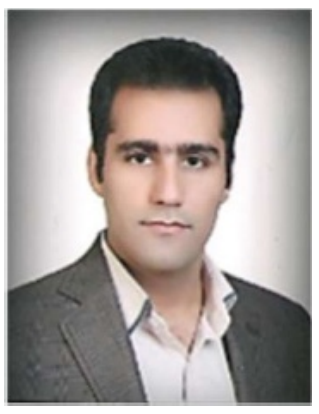

**مجید هارونی** تحصیلات خود را در مقطع کارشناسی مهندسی کامپیوتر در سال ۱۳۸۲ در دانشگاه آزاد اسلامی واحد نجف آباد به پایان رساند. مقاطع کارشناسی ارشد و دکترای مهندسی کامپیوتر را در دانشگاه تکنولوژی مالزی تحت بورسیه وزارت علوم، فناوری و نوآوری مالزی به‌ترتیب در سال‌های ۱۳۸۸ و ۱۳۹۲ تکمیل کرد. ایشان در حال حاضر عضو هیئت علمی واحد دولت‌آباد و سرپرست آزمایشگاه پردازش چندرسانه‌ای ادراکی این واحد دانشگاهی است. پژوهش‌های مورد علاقه ایشان در زمینه پردازش چندرسانه‌ای ادراکی، بازشناسی الگو و بینایی ماشین است.

نشانی رایانامهٔ ایشان عبارت‌است‌از:

**m.harouni@iauda.ac.ir**
**majid.harouni@gmail.com**

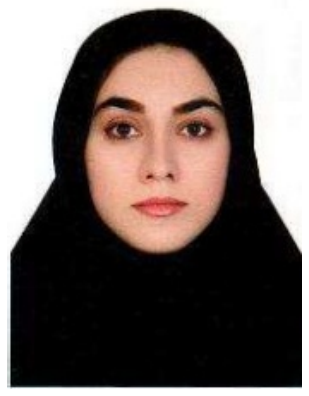

**مینا محمدی دشتی** در سال ۱۳۹۲ مقطع کارشناسی خود را در رشته مهندسی کامپیوتر در دانشگاه آزاد اسلامی واحد دولت آباد به پایان رسانده است. در سال ۱۳۹۵ مدرک کارشناسی ارشد خود را در همان رشته (گرایش معماری سیستم‌های کامپیوتری) از دانشگاه آزاد اسلامی واحد نجف آباد اخذ کرده است. پژوهش‌های مورد علاقه ایشان در زمینه پردازش تصویر و بینایی ماشین است.

نشانی رایانامهٔ ایشان عبارت‌است‌از:

**m.mohammadi96@yahoo.com**